\pdfoutput=1
\documentclass[11pt]{article}
\usepackage[final]{acl}

\title{
AgentDrug: Utilizing Large Language Models in An Agentic Workflow for Zero-Shot Molecular Editing
}
\author{
Khiem Le, Ting Hua, Nitesh V. Chawla\\
University of Notre Dame, IN, USA\\
\texttt{\{kle3, thua, nchawla\}@nd.edu}
}

\begin{document}
\maketitle

\begin{abstract}
Molecular editing—modifying a given molecule to improve desired properties—is a fundamental task in drug discovery. 
While LLMs hold the potential to solve this task using natural language to drive the editing, straightforward prompting achieves limited accuracy. 
In this work, we propose AgentDrug \footnote{\href{https://github.com/lhkhiem28/AgentDrug}{https://github.com/lhkhiem28/AgentDrug}}, an agentic workflow that leverages LLMs in a structured refinement process to achieve significantly higher accuracy.
AgentDrug defines a nested refinement loop: the inner loop uses feedback from cheminformatics toolkits to validate molecular structures, while the outer loop guides the LLM with generic feedback and a gradient-based objective to steer the molecule toward property improvement. 
We evaluate AgentDrug on benchmarks with both single- and multi-property editing under loose and strict thresholds. Results demonstrate significant performance gains over previous methods. 
With Qwen-2.5-3B, AgentDrug improves accuracy by 20.7\% (loose) and 16.8\% (strict) on six single-property tasks, and by 7.0\% and 5.3\% on eight multi-property tasks. 
With larger model Qwen-2.5-7B, AgentDrug further improves accuracy on 6 single-property objectives by 28.9\% (loose) and 29.0\% (strict), and on 8 multi-property objectives by 14.9\% (loose) and 13.2\% (strict).
\end{abstract}

\section{Introduction}
The process of drug discovery, which involves identifying molecules that can safely treat or influence a disease, is expensive and typically takes over a decade to result in an approved drug \cite{bateman2022563,sertkaya2024}. Recent advancements in AI for molecular studies have been revolutionizing this process by scaling many key tasks efficiently \cite{mak2019artificial}.
While significant progress has been made in molecular property prediction, synthesis, and retrosynthesis \cite{zhao2023gimlet, liu-etal-2024-reactxt}, molecular editing—a crucial step in refining candidate molecules—remains relatively underexplored.

\begin{figure}[!t]
\centering
\captionsetup[subfigure]{labelformat=empty}

\begin{subfigure}[t]{00.5\linewidth}
    \centering
    \includegraphics[height=0.9745\linewidth]{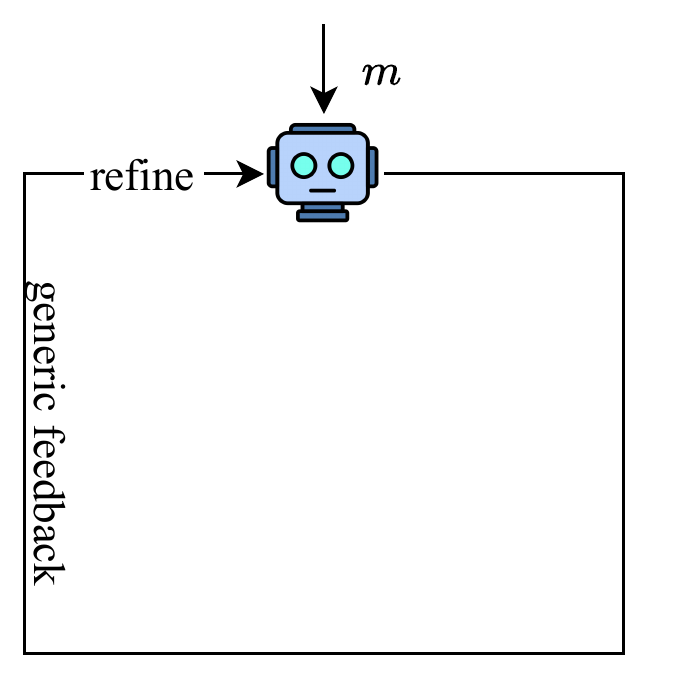}
    \caption{ChatDrug}
\end{subfigure}%
\hfill
\begin{subfigure}[t]{00.5\linewidth}
    \centering
    \includegraphics[height=0.9745\linewidth]{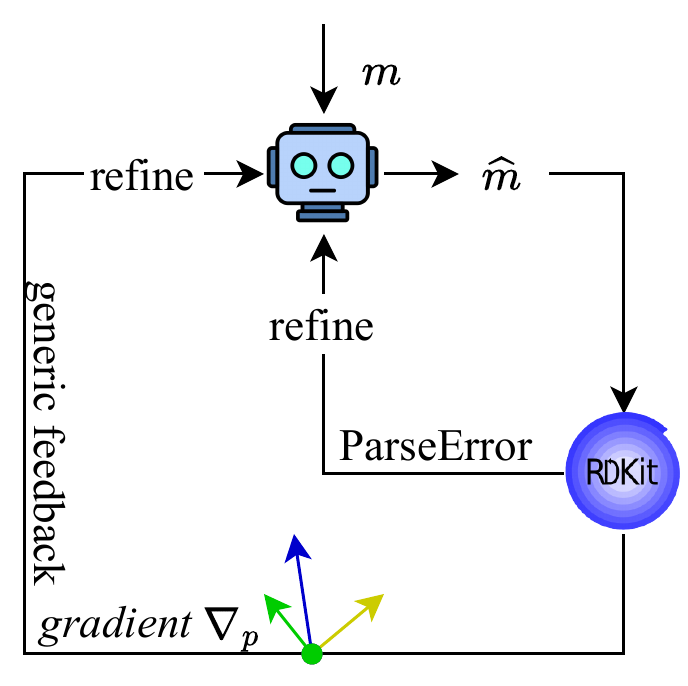}
    \caption{AgentDrug}
\end{subfigure}%
\hfill

\caption{An illustration of AgentDrug workflow}
\label{Figure-1}
\end{figure}

Molecular editing involves modifying a given molecule to improve specific desired properties. A key challenge in this task is the similarity constraint—the modified molecule must remain structurally similar to the original \cite{jorgensen2009efficient}, which differentiates it from de novo molecular generation \cite{schneider2005computer}. In practice, this task is routinely performed by chemists through manual iterations \cite{hoffer2018integrated}, making it labor-intensive and difficult to scale. Recent breakthroughs in large language models (LLMs) offer a promising alternative: LLMs can guide molecular editing via natural language, enabling more scalable and automated solutions. Unlike traditional supervised learning approaches \cite{he2021molecular, he2022transformer}, LLMs can reduce the dependence on labeled molecule pairs \cite{liu2023pre}, and exhibit zero-shot and open-vocabulary generalization beyond pre-defined objective sets. 
Moreover, they can also substantially reduce training costs.

Early attempts explore the use of LLMs for molecular editing through straightforward prompting, but report limited accuracy \cite{zhang2024future}. 
Building on the refinement strategies commonly employed by chemists, \citet{liu2024conversational} propose ChatDrug, a conversational framework that guides LLMs in an iterative refinement loop. This loop relies on generic feedback—for example, ``The modified molecule does not meet the objective''—to prompt further revisions. 
However, this setup has key limitations: the generic nature of the feedback causes LLMs to refine molecules merely toward the objective, and the method does not address molecular hallucination \cite{guo2023can}, where LLMs produce invalid molecular structures. Consequently, the retrieval step is often disabled due to the high frequency of invalid outputs, diminishing its effectiveness.

In this work, we propose AgentDrug  (Figure~\ref{Figure-1}), an agentic workflow that addresses the limitations of previous work. AgentDrug designed  a nested refinement loop to systematically improve both the validity and quality of the modified molecule through iterative interactions between the LLM and external tools. The inner loop interacts with a cheminformatics toolkit to detect and extract ParseError messages, which serve as feedback to help the LLM revise the modified molecule into a valid one. Once validity is achieved, i.e., the molecule passes the inner loop, the outer loop provides both generic feedback and an explicit gradient signal to guide the LLM in editing the molecule toward the target objective. 
In addition, AgentDrug retrieves molecules from a prepared database based on two criteria: (1) similarity to the modified molecule and (2) satisfaction of the editing objective. These molecules serve as in-context examples to guide the LLM. 
With the explicit \textit{gradient}, the LLM gains actionable guidance analogous to \textit{gradient ascent}, enabling more effective refinement. 
In contrast to these earlier methods, AgentDrug enforces molecular validity early in its workflow, allowing the retrieval step to function more reliably and play a more impactful role in guiding editing.

\section{Methodology}

Formally, molecular editing is the task of modifying a given molecule $m$, expressed in the SMILES string \cite{weininger1988smiles}. The goal is to adjust a set of desired molecular properties $p = \{p_i\}_{i=1}^N$ by thresholds $d = \{d_i\}_{i=1}^N$. Each threshold $d_i \in \mathbb{R}$ is associated with a direction $\sigma(d_i)$, either \texttt{+} (indicating an increase) or \texttt{-} (indicating a decrease), which specifies whether the property $p_i$ should be improved or reduced. Given an LLM $\mathcal{M}$, the editing is defined as:
\begin{equation}
\label{eqn-obj}
\begin{aligned}
    \widehat{m} = \mathcal{M}(p || d || m) \myspace \text{s.t.} \myspace \prod_{i=1}^{N} 
    \underset{p_i, d_i}{E} (\widehat{m}) = 1 \myspace \text{where}\nonumber\\
    \underset{p_i, d_i}{E} (\widehat{m}) = \mathbbm{1}[\sigma(d_i)((p_i[\widehat{m}] - p_i[m]) - d_i) \ge 0].
\end{aligned}
\end{equation}
Here, $||$ denotes concatenation, and $\mathbbm{1} \in {0, 1}$ is an indicator function that evaluates whether the change in property $p_i$ satisfies the specified threshold $d_i$ in the intended direction.

\begin{table}[!t]
    \centering
    \caption{Definition of the six categories of ParseError. }
    \label{tab-ParseError}
    \small
    \begin{tabularx}{\linewidth}{lX}
    \midrule
    ParseError      & Definition 
    \\\midrule
    syntax          & The SMILES string does not follow the correct SMILES grammar, often due to unrecognized characters or patterns. 
    \\\midrule
    parentheses     & The SMILES string contains unmatched parentheses, disrupting branching logic. 
    \\\midrule
    duplicate bond  & The SMILES string contains a bond that is defined more than once between the same pair of atoms. 
    \\\midrule
    valence         & The SMILES string contains an atom that is assigned more bonds than its allowed valence. 
    \\\midrule
    aromaticity     & The SMILES string contains misused aromatic atoms, e.g., marking a non-ring atom as aromatic or causing kekulization conflicts. 
    \\\midrule
    unclosed ring   & The SMILES string contains a ring closure digit that appears only once, meaning the ring was opened but not closed. 
    \\\midrule
    \end{tabularx}
    \end{table}

AgentDrug begins by utilizing LLM $\mathcal{M}$ to generate an initial modified molecule $\widehat{m}$. However, LLMs often suffer from the ``molecule hallucination'' phenomenon, where the generated molecule is chemically invalid. To address this, AgentDrug incorporates a validation and refinement loop inspired by debugging practices.

Specifically, the validity of $\widehat{m}$ is checked using RDKit \cite{landrum2013rdkit}, a cheminformatics toolkit that parses the SMILES string and returns a ParseError if the molecule is invalid. There are six categories of ParseError, as summarized in Table~\ref{tab-ParseError}. When a ParseError is detected, it is provided as feedback to the LLM, prompting it to iteratively refine $\widehat{m}$ until a valid molecule is produced.

\begin{table*}[!ht]
    \centering
    \caption{The results (T = 3) on a set of single- and multi-property objectives with loose and strict thresholds. }
    \label{tab-2}
    \tiny
    \setlength\tabcolsep{0pt}
    \begin{tabular*}{\linewidth}{@{\extracolsep{\fill}} llcccccccccccccccccccc }
    %%%%%%%%%%%%%%%%%%%%%%%%%%%%%%%%%%%%%%%%%%%%%%%%%%%%%%%%%%%%%%%%%%%%%%%%%%%%

    %%%%%%%%%%%%%%%%%%%%%%%%%%%%%%%%%%%%%%%%%%%%%%%%%%%%%
    \midrule
    \multirow[t]{3}{*}{$p$} &\multirow[t]{3}{*}{$d$} &\multicolumn{10}{c}{Qwen2.5-3B} &\multicolumn{10}{c}{Qwen2.5-7B} \\
    & &\multicolumn{2}{c}{vanilla} &\multicolumn{2}{c}{REINVENT} &\multicolumn{2}{c}{ChatDrug} &\multicolumn{2}{c}{AgentDrug} &\multicolumn{2}{c}{AgentDrug$\dag$} &\multicolumn{2}{c}{vanilla} &\multicolumn{2}{c}{REINVENT} &\multicolumn{2}{c}{ChatDrug} &\multicolumn{2}{c}{AgentDrug} &\multicolumn{2}{c}{AgentDrug$\dag$} \\
    & &valid &accuracy &valid &accuracy &valid &accuracy &valid &accuracy &valid &accuracy &valid &accuracy &valid &accuracy &valid &accuracy &valid &accuracy &valid &accuracy \\
    \midrule
    \multirow[t]{2}{*}{\texttt{+}LogP}                 &\texttt{l}            &\ul{68.7} &21.3 &66.0 &17.5 &65.8 &27.6 &\textbf{74.1} &\textbf{43.0} &68.0 &\ul{32.7} &\ul{76.3} &29.8 &75.8 &29.5 &72.5 &40.2 &\textbf{87.0} &\textbf{71.7} &73.3 &\ul{46.2} \\
                                                       &\texttt{s}            &66.0 &11.2 &\ul{69.4} &11.5 &63.5 &\ul{22.5} &\textbf{72.5} &\textbf{37.7} &62.9 &22.0 &\ul{74.3} &16.7 &74.1 &16.7 &70.4 &33.1 &\textbf{84.4} &\textbf{66.2} &73.5 &\ul{45.6} \\
    \multirow[t]{2}{*}{\texttt{-}LogP}                 &\texttt{l}            &69.0 &12.4 &\ul{71.4} &16.1 &69.0 &26.6 &\textbf{79.0} &\textbf{49.8} &66.4 &\ul{28.2} &\ul{77.8} &28.8 &77.2 &27.8 &73.5 &41.2 &\textbf{85.8} &\textbf{70.8} &74.9 &\ul{49.1} \\
                                                       &\texttt{s}            &70.9 &09.9 &\ul{73.3} &05.5 &67.1 &\ul{20.5} &\textbf{75.8} &\textbf{40.5} &61.9 &16.1 &74.6 &11.6 &74.6 &11.6 &73.5 &39.3 &\textbf{83.0} &\textbf{63.5} &\ul{77.8} &\ul{55.6} \\
    \multirow[t]{2}{*}{\texttt{+}TPSA}                 &\texttt{l}            &68.5 &13.7 &\ul{70.2} &15.4 &63.5 &21.0 &\textbf{76.6} &\textbf{46.4} &65.4 &\ul{26.1} &\ul{72.5} &23.2 &72.5 &23.2 &\ul{72.5} &37.7 &\textbf{79.4} &\textbf{57.9} &70.4 &\ul{40.8} \\
                                                       &\texttt{s}            &\ul{67.3} &08.8 &67.1 &09.4 &63.7 &\ul{20.7} &\textbf{72.5} &\textbf{36.6} &61.7 &18.8 &\ul{69.0} &20.3 &69.0 &20.3 &66.0 &27.4 &\textbf{76.0} &\textbf{51.0} &67.8 &\ul{35.6} \\
    \multirow[t]{2}{*}{\texttt{-}TPSA}                 &\texttt{l}            &68.5 &10.6 &\ul{69.2} &11.1 &66.9 &\ul{23.7} &\textbf{74.3} &\textbf{39.0} &64.1 &21.8 &73.8 &11.8 &73.5 &11.4 &73.5 &36.8 &\textbf{86.2} &\textbf{69.8} &\ul{77.2} &\ul{52.5} \\
                                                       &\texttt{s}            &69.2 &06.2 &\ul{71.7} &07.2 &65.8 &22.7 &\textbf{76.9} &\textbf{38.8} &65.4 &\ul{23.9} &72.5 &04.4 &71.9 &04.3 &70.7 &31.5 &\textbf{82.3} &\textbf{63.4} &\ul{73.8} &\ul{45.8} \\
    \multirow[t]{2}{*}{\texttt{+}QED}                  &\texttt{l}            &68.7 &11.3 &70.2 &13.3 &\ul{70.2} &29.1 &\textbf{78.7} &\textbf{48.8} &68.5 &\ul{31.5} &76.6 &18.8 &76.3 &18.3 &71.7 &33.0 &\textbf{83.7} &\textbf{65.7} &\ul{77.5} &\ul{54.7} \\
                                                       &\texttt{s}            &\ul{70.7} &02.5 &70.2 &04.2 &66.2 &10.9 &\textbf{73.5} &\textbf{23.2} &64.7 &\ul{12.3} &\ul{79.0} &05.1 &78.4 &04.3 &74.1 &20.4 &\textbf{80.0} &\textbf{48.0} &78.1 &\ul{41.0} \\
    \multirow[t]{2}{*}{\texttt{-}QED}                  &\texttt{l}            &\ul{73.5} &23.9 &72.2 &22.0 &68.0 &27.2 &\textbf{80.0} &\textbf{52.4} &68.0 &\ul{34.4} &76.9 &28.5 &76.6 &28.4 &76.9 &49.6 &\textbf{88.5} &\textbf{76.5} &\ul{78.7} &\ul{57.5} \\
                                                       &\texttt{s}            &\ul{71.2} &08.2 &69.9 &05.9 &66.7 &16.0 &\textbf{75.2} &\textbf{37.6} &66.4 &\ul{26.9} &\ul{77.5} &13.2 &77.2 &13.1 &71.4 &33.9 &\textbf{84.4} &\textbf{67.5} &74.3 &\ul{47.2} \\
    \multirow[t]{2}{*}{Average}                        &\texttt{l}            &69.5 &15.5 &\ul{69.9} &15.9 &67.2 &25.9 &\textbf{77.1} &\textbf{46.6} &66.7 &\ul{29.1} &\ul{75.6} &23.5 &75.3 &23.1 &73.4 &39.8 &\textbf{85.1} &\textbf{68.7} &75.3 &\ul{50.1} \\
                                                       &\texttt{s}            &69.2 &07.8 &\ul{70.3} &07.3 &65.5 &18.9 &\textbf{74.4} &\textbf{35.7} &63.8 &\ul{20.0} &\ul{74.5} &11.9 &74.2 &11.7 &71.0 &30.9 &\textbf{81.7} &\textbf{59.9} &74.2 &\ul{45.1} \\
    \midrule
    \multirow[t]{2}{*}{\texttt{+}LogP \texttt{+}TPSA}  &\texttt{l} \texttt{l} &\ul{67.1} &03.4 &66.4 &06.3 &66.2 &16.9 &\textbf{69.0} &\textbf{25.5} &63.9 &\ul{18.2} &\ul{72.7} &05.1 &72.2 &05.0 &69.7 &26.8 &\textbf{78.1} &\textbf{49.2} &\ul{72.7} &\ul{43.6} \\
                                                       &\texttt{s} \texttt{s} &68.0 &03.4 &\ul{68.0} &04.4 &65.6 &\ul{18.7} &\textbf{69.4} &\textbf{24.3} &63.3 &17.7 &\ul{71.2} &04.6 &70.9 &04.6 &66.7 &22.0 &\textbf{81.3} &\textbf{53.3} &69.9 &\ul{37.8} \\
    \multirow[t]{2}{*}{\texttt{+}LogP \texttt{-}TPSA}  &\texttt{l} \texttt{l} &\ul{69.0} &04.8 &67.6 &05.7 &64.9 &10.1 &\textbf{69.9} &\textbf{14.7} &64.5 &\ul{13.6} &\ul{79.0} &09.5 &78.4 &09.4 &74.3 &18.6 &\textbf{81.3} &\textbf{27.2} &76.9 &\ul{22.3} \\
                                                       &\texttt{s} \texttt{s} &69.7 &03.1 &\ul{69.9} &03.5 &62.7 &05.3 &\textbf{70.9} &\textbf{08.2} &62.5 &\ul{06.6} &72.7 &01.8 &71.9 &01.8 &73.0 &10.6 &\textbf{80.3} &\textbf{16.1} &\ul{73.8} &\ul{12.9} \\
    \multirow[t]{2}{*}{\texttt{-}LogP \texttt{+}TPSA}  &\texttt{l} \texttt{l} &\ul{69.0} &04.8 &67.8 &07.5 &68.0 &16.7 &\textbf{69.9} &\textbf{25.9} &65.6 &\ul{19.3} &78.7 &26.0 &\ul{78.7} &26.0 &72.7 &39.6 &\textbf{80.0} &\textbf{56.0} &76.9 &\ul{52.3} \\
                                                       &\texttt{s} \texttt{s} &67.3 &02.7 &\ul{70.7} &03.5 &63.9 &10.5 &\textbf{71.9} &\textbf{23.7} &66.7 &\ul{18.0} &76.6 &13.4 &\ul{76.6} &13.4 &67.8 &25.8 &\textbf{77.5} &\textbf{50.4} &72.5 &\ul{42.0} \\
    \multirow[t]{2}{*}{\texttt{-}LogP \texttt{-}TPSA}  &\texttt{l} \texttt{l} &66.9 &03.3 &\textbf{70.4} &03.2 &66.7 &04.3 &\ul{70.4} &\textbf{10.2} &63.7 &\ul{04.8} &75.8 &03.0 &75.8 &03.0 &72.7 &05.1 &\textbf{84.0} &\textbf{14.7} &\ul{77.5} &\ul{12.0} \\
                                                       &\texttt{s} \texttt{s} &67.8 &00.3 &\ul{70.4} &00.0 &64.5 &\ul{01.3} &\textbf{74.9} &\textbf{02.3} &66.7 &\ul{01.3} &76.6 &00.4 &76.3 &00.4 &\ul{77.2} &02.3 &\textbf{83.7} &\textbf{07.1} &\ul{77.2} &\ul{04.3} \\
    \multirow[t]{2}{*}{\texttt{+}LogP \texttt{+}QED}   &\texttt{l} \texttt{l} &66.2 &07.6 &\ul{69.4} &09.0 &65.1 &12.1 &\textbf{71.4} &\textbf{25.0} &65.1 &\ul{18.6} &74.9 &06.4 &74.3 &06.3 &73.8 &31.0 &\textbf{79.7} &\textbf{55.8} &\ul{76.3} &\ul{50.8} \\
                                                       &\texttt{s} \texttt{s} &66.9 &01.0 &\textbf{68.0} &01.0 &65.6 &06.2 &\ul{68.0} &\textbf{16.3} &65.8 &\ul{09.2} &72.5 &01.1 &72.2 &01.1 &71.9 &14.0 &\textbf{78.1} &\textbf{38.3} &\ul{74.1} &\ul{33.0} \\
    \multirow[t]{2}{*}{\texttt{+}LogP \texttt{-}QED}   &\texttt{l} \texttt{l} &\ul{71.2} &15.3 &68.0 &11.9 &66.2 &16.6 &\textbf{71.9} &\textbf{20.5} &66.9 &\ul{16.7} &\ul{75.5} &13.2 &75.2 &13.2 &73.8 &\ul{20.3} &\textbf{84.7} &\textbf{25.4} &75.2 &16.5 \\
                                                       &\texttt{s} \texttt{s} &67.6 &04.1 &\textbf{69.4} &04.5 &63.1 &04.7 &\ul{68.0} &\textbf{09.5} &62.9 &\ul{06.6} &74.1 &\textbf{08.2} &73.8 &\ul{08.1} &71.4 &03.9 &\textbf{79.0} &06.3 &\ul{74.6} &05.2 \\
    \multirow[t]{2}{*}{\texttt{-}LogP \texttt{+}QED}   &\texttt{l} \texttt{l} &67.6 &03.4 &\ul{70.4} &04.9 &66.7 &13.3 &\textbf{71.4} &\textbf{23.9} &66.4 &\ul{17.6} &\ul{76.9} &10.0 &76.3 &09.2 &74.1 &33.0 &\textbf{84.0} &\textbf{61.3} &76.6 &\ul{51.3} \\
                                                       &\texttt{s} \texttt{s} &69.2 &01.4 &\textbf{71.4} &01.1 &65.4 &\ul{07.2} &\ul{71.2} &\textbf{10.7} &67.6 &06.8 &\ul{78.4} &00.4 &78.1 &00.0 &74.6 &17.9 &77.8 &\ul{31.9} &\textbf{80.0} &\textbf{36.4} \\
    \multirow[t]{2}{*}{\texttt{-}LogP \texttt{-}QED}   &\texttt{l} \texttt{l} &\textbf{71.9} &08.6 &\ul{71.4} &09.3 &67.1 &\ul{09.7} &71.4 &\textbf{10.4} &67.1 &09.1 &\ul{79.4} &19.1 &78.7 &18.9 &75.8 &\ul{19.7} &\textbf{85.1} &\textbf{24.3} &\ul{79.4} &19.1 \\
                                                       &\texttt{s} \texttt{s} &\textbf{71.7} &\textbf{03.6} &\ul{71.4} &01.1 &65.8 &01.0 &69.0 &02.4 &66.4 &\ul{03.3} &\ul{77.2} &03.5 &76.9 &03.5 &74.3 &\textbf{05.6} &\textbf{81.0} &04.9 &76.9 &\ul{05.0} \\
    \multirow[t]{2}{*}{Average}                        &\texttt{l} \texttt{l} &68.6 &06.4 &\ul{68.9} &07.2 &66.4 &12.5 &\textbf{70.7} &\textbf{19.5} &65.4 &\ul{14.7} &\ul{76.6} &11.5 &76.2 &11.4 &73.4 &24.3 &\textbf{82.1} &\textbf{39.2} &76.4 &\ul{33.5} \\
                                                       &\texttt{s} \texttt{s} &68.5 &02.5 &\ul{69.9} &02.4 &64.6 &06.9 &\textbf{70.4} &\textbf{12.2} &65.2 &\ul{08.7} &\ul{74.9} &04.2 &74.6 &04.1 &72.1 &12.8 &\textbf{79.8} &\textbf{26.0} &74.9 &\ul{22.1} \\
    \midrule
    %%%%%%%%%%%%%%%%%%%%%%%%%%%%%%%%%%%%%%%%%%%%%%%%%%%%%
    %%%%%%%%%%%%%%%%%%%%%%%%%%%%%%%%%%%%%%%%%%%%%%%%%%%%%
    \midrule
    \multirow[t]{3}{*}{$p$} &\multirow[t]{3}{*}{$d$} &\multicolumn{10}{c}{Llama-3.1-8B} &\multicolumn{10}{c}{Llama-3.1-70B} \\
    & &\multicolumn{2}{c}{vanilla} &\multicolumn{2}{c}{REINVENT} &\multicolumn{2}{c}{ChatDrug} &\multicolumn{2}{c}{AgentDrug} &\multicolumn{2}{c}{AgentDrug$\dag$} &\multicolumn{2}{c}{vanilla} &\multicolumn{2}{c}{REINVENT} &\multicolumn{2}{c}{ChatDrug} &\multicolumn{2}{c}{AgentDrug} &\multicolumn{2}{c}{AgentDrug$\dag$} \\
    & &valid &accuracy &valid &accuracy &valid &accuracy &valid &accuracy &valid &accuracy &valid &accuracy &valid &accuracy &valid &accuracy &valid &accuracy &valid &accuracy \\
    \midrule
    \multirow[t]{2}{*}{\texttt{+}LogP}                 &\texttt{l}            &76.3 &40.1 &73.3 &32.9 &76.0 &48.3 &\textbf{85.8} &\textbf{69.1} &\ul{76.6} &\ul{52.9} &82.0 &57.4 &81.4 &56.9 &\ul{83.7} &\ul{67.4} &\textbf{94.8} &\textbf{89.1} &82.6 &65.3 \\
                                                       &\texttt{s}            &71.7 &26.2 &\ul{75.4} &26.8 &71.4 &38.6 &\textbf{81.6} &\textbf{61.6} &71.9 &\ul{41.4} &\ul{87.7} &35.1 &87.4 &35.0 &80.3 &59.8 &\textbf{92.2} &\textbf{81.6} &83.3 &\ul{65.8} \\
    \multirow[t]{2}{*}{\texttt{-}LogP}                 &\texttt{l}            &82.0 &25.8 &\textbf{84.8} &33.5 &75.8 &36.4 &\ul{83.0} &\textbf{56.8} &74.9 &\ul{43.4} &\ul{82.6} &55.0 &82.0 &53.0 &78.4 &56.1 &\textbf{91.7} &\textbf{83.5} &79.0 &\ul{57.7} \\
                                                       &\texttt{s}            &79.0 &13.4 &\ul{81.7} &07.5 &72.7 &27.3 &\textbf{87.0} &\textbf{60.9} &72.2 &\ul{36.8} &82.0 &30.3 &\ul{82.0} &30.2 &77.8 &52.9 &\textbf{94.8} &\textbf{84.8} &79.0 &\ul{56.5} \\
    \multirow[t]{2}{*}{\texttt{+}TPSA}                 &\texttt{l}            &75.2 &25.9 &\ul{77.0} &29.2 &74.6 &42.5 &\textbf{85.8} &\textbf{70.4} &72.5 &\ul{43.8} &\ul{80.6} &59.7 &80.6 &59.6 &80.0 &59.6 &\textbf{90.1} &\textbf{80.2} &80.0 &\ul{60.0} \\
                                                       &\texttt{s}            &\ul{74.9} &24.0 &74.7 &25.6 &74.6 &44.0 &\textbf{85.8} &\textbf{70.0} &73.8 &\ul{46.9} &79.0 &55.3 &79.0 &55.4 &79.4 &58.7 &\textbf{93.9} &\textbf{86.8} &\ul{82.0} &\ul{63.9} \\
    \multirow[t]{2}{*}{\texttt{-}TPSA}                 &\texttt{l}            &78.4 &16.5 &\ul{79.2} &17.2 &74.9 &41.6 &\textbf{89.7} &\textbf{75.8} &76.3 &\ul{49.6} &\ul{85.1} &37.0 &84.8 &35.8 &80.0 &59.2 &\textbf{96.6} &\textbf{90.8} &84.7 &\ul{68.6} \\
                                                       &\texttt{s}            &82.0 &17.2 &\ul{84.9} &19.9 &77.2 &40.9 &\textbf{86.6} &\textbf{66.7} &74.1 &\ul{45.9} &\ul{86.6} &31.6 &85.9 &31.0 &77.5 &53.1 &\textbf{95.2} &\textbf{89.5} &83.7 &\ul{66.1} \\
    \multirow[t]{2}{*}{\texttt{+}QED}                  &\texttt{l}            &80.0 &18.4 &\ul{81.7} &21.7 &75.2 &39.9 &\textbf{85.1} &\textbf{60.4} &74.9 &\ul{43.8} &\ul{86.6} &34.6 &86.3 &33.7 &81.3 &52.8 &\textbf{93.5} &\textbf{82.7} &83.3 &\ul{62.5} \\
                                                       &\texttt{s}            &\ul{80.6} &02.8 &80.0 &04.7 &74.6 &17.9 &\textbf{83.0} &\textbf{45.6} &75.8 &\ul{28.0} &\ul{90.1} &06.3 &89.4 &05.3 &85.1 &36.6 &\textbf{94.3} &\textbf{62.7} &84.7 &\ul{45.8} \\
    \multirow[t]{2}{*}{\texttt{-}QED}                  &\texttt{l}            &\ul{77.5} &31.8 &76.1 &29.3 &73.0 &39.4 &\textbf{86.6} &\textbf{68.8} &75.5 &\ul{49.8} &\ul{80.6} &41.5 &80.3 &41.3 &77.8 &55.3 &\textbf{94.3} &\textbf{87.3} &79.4 &\ul{58.7} \\
                                                       &\texttt{s}            &\ul{81.6} &22.4 &80.2 &16.3 &71.9 &34.2 &\textbf{85.5} &\textbf{66.2} &69.7 &\ul{34.8} &\ul{87.0} &06.1 &86.6 &06.1 &78.7 &52.4 &\textbf{92.6} &\textbf{80.1} &81.0 &\ul{58.3} \\
    \multirow[t]{2}{*}{Average}                        &\texttt{l}            &78.2 &26.4 &\ul{78.7} &27.0 &74.9 &41.3 &\textbf{86.0} &\textbf{66.9} &75.1 &\ul{47.2} &\ul{82.9} &47.5 &82.6 &46.8 &80.2 &58.4 &\textbf{93.5} &\textbf{85.6} &81.5 &\ul{62.1} \\
                                                       &\texttt{s}            &78.3 &17.7 &\ul{79.5} &16.5 &73.8 &33.8 &\textbf{84.9} &\textbf{61.8} &72.9 &\ul{39.0} &\ul{85.4} &27.5 &85.1 &27.1 &79.8 &52.3 &\textbf{93.8} &\textbf{80.9} &82.3 &\ul{59.4} \\
    \midrule
    \multirow[t]{2}{*}{\texttt{+}LogP \texttt{+}TPSA}  &\texttt{l} \texttt{l} &\ul{73.3} &11.0 &72.5 &20.4 &67.1 &23.8 &\textbf{73.8} &\textbf{42.4} &69.7 &\ul{33.5} &\ul{84.7} &17.4 &84.1 &17.2 &74.3 &36.4 &\textbf{88.1} &\textbf{55.9} &81.0 &\ul{45.3} \\
                                                       &\texttt{s} \texttt{s} &69.2 &10.0 &69.2 &13.0 &66.4 &19.6 &\textbf{71.7} &\ul{31.9} &\ul{70.2} &\textbf{32.3} &\ul{84.4} &10.6 &84.0 &10.6 &76.3 &34.7 &\textbf{87.0} &\textbf{57.0} &82.0 &\ul{48.8} \\
    \multirow[t]{2}{*}{\texttt{+}LogP \texttt{-}TPSA}  &\texttt{l} \texttt{l} &\ul{76.3} &11.8 &74.7 &14.1 &71.2 &19.6 &\textbf{78.4} &\textbf{27.1} &72.5 &\ul{24.6} &\ul{85.8} &41.6 &85.2 &41.2 &83.7 &\ul{48.1} &\textbf{91.3} &\textbf{58.9} &83.3 &47.5 \\
                                                       &\texttt{s} \texttt{s} &74.6 &06.0 &\ul{74.9} &06.7 &70.4 &08.8 &\textbf{80.6} &\textbf{16.1} &70.7 &\ul{11.3} &\textbf{91.3} &21.5 &90.4 &21.5 &79.7 &\ul{30.3} &\textbf{91.3} &\textbf{33.3} &83.7 &28.4 \\
    \multirow[t]{2}{*}{\texttt{-}LogP \texttt{+}TPSA}  &\texttt{l} \texttt{l} &\ul{73.8} &11.4 &72.5 &17.8 &71.2 &19.2 &\textbf{78.4} &\textbf{42.0} &70.2 &\ul{31.9} &83.0 &57.7 &83.0 &57.6 &79.4 &57.1 &\textbf{91.3} &\textbf{80.4} &\ul{84.0} &\ul{65.5} \\
                                                       &\texttt{s} \texttt{s} &70.4 &06.7 &\ul{73.9} &08.7 &70.9 &14.5 &\textbf{75.5} &\ul{29.8} &70.7 &\textbf{30.4} &87.0 &37.0 &\ul{87.0} &37.0 &78.7 &53.5 &\textbf{92.6} &\textbf{74.1} &84.7 &\ul{57.2} \\
    \multirow[t]{2}{*}{\texttt{-}LogP \texttt{-}TPSA}  &\texttt{l} \texttt{l} &77.5 &03.1 &\textbf{81.6} &03.0 &71.9 &\textbf{09.7} &\ul{79.7} &\ul{09.6} &74.9 &08.2 &\ul{81.0} &02.8 &80.9 &02.9 &70.2 &05.3 &\textbf{85.8} &\textbf{09.4} &74.9 &\ul{06.7} \\
                                                       &\texttt{s} \texttt{s} &76.3 &00.8 &\ul{79.3} &00.0 &73.0 &05.1 &\textbf{80.6} &\textbf{06.0} &75.2 &\ul{05.3} &\ul{81.3} &02.0 &81.0 &01.9 &75.2 &\textbf{06.4} &\textbf{88.9} &\ul{06.2} &77.8 &05.1 \\
    \multirow[t]{2}{*}{\texttt{+}LogP \texttt{+}QED}   &\texttt{l} \texttt{l} &81.0 &13.4 &\textbf{84.9} &15.9 &73.0 &20.4 &\ul{83.0} &\textbf{46.0} &71.9 &\ul{29.1} &\ul{84.4} &20.7 &83.7 &20.4 &78.4 &35.7 &\textbf{87.7} &\textbf{47.8} &78.7 &\ul{40.9} \\
                                                       &\texttt{s} \texttt{s} &\ul{80.0} &03.2 &\textbf{81.3} &03.3 &71.2 &07.5 &74.6 &\textbf{23.9} &74.1 &\ul{18.5} &\textbf{89.3} &02.2 &\ul{88.9} &02.2 &82.0 &20.1 &88.5 &\textbf{36.3} &85.5 &\ul{26.9} \\
    \multirow[t]{2}{*}{\texttt{+}LogP \texttt{-}QED}   &\texttt{l} \texttt{l} &\ul{76.0} &25.9 &72.6 &20.1 &74.9 &\ul{28.5} &\textbf{77.2} &\textbf{35.5} &71.4 &24.3 &\ul{84.7} &53.0 &84.4 &52.8 &82.3 &53.5 &\textbf{91.3} &\textbf{65.3} &\ul{84.7} &\ul{55.5} \\
                                                       &\texttt{s} \texttt{s} &\ul{86.2} &15.5 &\textbf{88.5} &\ul{17.1} &81.0 &\textbf{22.7} &75.8 &16.3 &76.6 &15.3 &\ul{89.7} &12.6 &89.3 &12.4 &85.1 &\textbf{21.3} &\textbf{90.9} &12.7 &88.9 &\ul{14.2} \\
    \multirow[t]{2}{*}{\texttt{-}LogP \texttt{+}QED}   &\texttt{l} \texttt{l} &\ul{78.4} &08.2 &\textbf{81.7} &11.9 &77.2 &21.6 &77.5 &\textbf{32.9} &75.2 &\ul{29.3} &\ul{80.6} &16.9 &80.0 &15.5 &74.6 &23.9 &\textbf{87.0} &\textbf{38.7} &79.0 &\ul{32.0} \\
                                                       &\texttt{s} \texttt{s} &77.8 &01.2 &\textbf{80.3} &00.9 &76.0 &07.2 &\ul{78.1} &\textbf{21.5} &77.8 &\ul{18.7} &\ul{85.5} &02.6 &85.2 &00.0 &81.0 &13.8 &\textbf{85.8} &\textbf{28.8} &84.4 &\ul{21.9} \\
    \multirow[t]{2}{*}{\texttt{-}LogP \texttt{-}QED}   &\texttt{l} \texttt{l} &\ul{77.8} &13.2 &77.3 &14.3 &75.5 &\ul{14.3} &\textbf{81.3} &\textbf{14.6} &73.5 &11.0 &\ul{78.4} &34.1 &77.8 &33.8 &75.5 &\ul{35.9} &\textbf{84.4} &\textbf{48.9} &74.9 &33.0 \\
                                                       &\texttt{s} \texttt{s} &\ul{83.0} &\ul{24.5} &82.7 &07.3 &78.1 &22.6 &\textbf{86.2} &\textbf{26.3} &81.6 &22.4 &\ul{84.0} &07.1 &83.7 &07.1 &73.3 &\ul{08.8} &\textbf{88.9} &\textbf{11.1} &80.6 &08.1 \\
    \multirow[t]{2}{*}{Average}                        &\texttt{l} \texttt{l} &76.8 &12.3 &\ul{77.1} &13.8 &72.7 &19.6 &\textbf{78.7} &\textbf{31.3} &72.4 &\ul{24.0} &\ul{82.8} &30.5 &82.4 &30.1 &77.3 &37.0 &\textbf{88.4} &\textbf{50.7} &80.1 &\ul{40.8} \\
                                                       &\texttt{s} \texttt{s} &77.2 &08.5 &\textbf{78.8} &08.3 &73.4 &13.5 &\ul{77.9} &\textbf{21.5} &74.6 &\ul{19.3} &\ul{86.5} &11.9 &86.2 &11.7 &78.9 &23.6 &\textbf{89.2} &\textbf{32.4} &83.4 &\ul{26.3} \\
    \midrule
    %%%%%%%%%%%%%%%%%%%%%%%%%%%%%%%%%%%%%%%%%%%%%%%%%%%%%

    %%%%%%%%%%%%%%%%%%%%%%%%%%%%%%%%%%%%%%%%%%%%%%%%%%%%%%%%%%%%%%%%%%%%%%%%%%%%
    \end{tabular*}
    \end{table*}

Once the modified molecule $\widehat{m}$ is valid (it passes the inner loop), AgentDrug generates generic feedback followed by an explicit \textit{gradient} signal $\nabla_p$ to guide further editing:

\begin{equation}
    \label{eqn-1}
    \nabla_p = \{ \sigma(d_i)|(p_i[\widehat{m}] - p_i[m]) - d_i| \}_{i=1}^N, 
\end{equation}

where $\sigma(d_i)$ denotes the desired direction of change (increase or decrease) for property $p_i$. The gradient encodes both the direction and the magnitude of required adjustments to the desired properties $p$, and is used to explicitly steer the LLM $\mathcal{M}$ in refining $\widehat{m}$ toward the editing objective.

Additionally, AgentDrug retrieves a molecule $m_e$ from a prepared database $\mathcal{D}$. This molecule is selected for its similarity to the modified molecule $\widehat{m}$ and its ability to meet the editing objective:
\begin{equation}
    \label{eqn-2}
    m_e = \underset{m_e' \in \mathcal{D}}{\text{argmax}} \left\langle m_e', \widehat{m} \right\rangle \wedge \prod_{i=1}^{N} \underset{p_i, d_i}{E} (m_e')
    \end{equation}
where $\left\langle m_e', \widehat{m} \right\rangle$ represents the Tanimoto similarity \cite{bajusz2015tanimoto} between $m_e'$ and $\widehat{m}$. The molecule $m_e$ serves as an in-context example, helping the LLM $\mathcal{M}$ refine $\widehat{m}$ toward the desired properties. By ensuring $\widehat{m}$ is valid, the retrieval step becomes more efficient and effective.

\section{Evaluation}

\begin{table}[!ht]
    \centering
    \caption{The results (T = 3) on a set of single- and multi-property objectives with loose and strict thresholds. }
    \label{tab-3}
    \tiny
    \setlength\tabcolsep{0pt}
    \begin{tabular*}{\linewidth}{@{\extracolsep{\fill}} llcccccc }
    %%%%%%%%%%%%%%%%%%%%%%%%%%%%%%%%%%%%%%%%%%%%%%%%%%%%%%%%%%%%%%

    %%%%%%%%%%%%%%%%%%%%%%%%%%%%%%%%%%%%%%%%%%%%%%%%%%%%%
    \midrule
    \multirow[t]{3}{*}{$p$} &\multirow[t]{3}{*}{$d$} &\multicolumn{3}{c}{Qwen2.5-3B} &\multicolumn{3}{c}{Qwen2.5-7B} \\
    & &ChatDrug &AgentDrug &AgentDrug$\dag$ &ChatDrug &AgentDrug &AgentDrug$\dag$ \\
    & &similarity &similarity &similarity &similarity &similarity &similarity \\
    \midrule
    \multirow[t]{2}{*}{\texttt{+}LogP}                 &\texttt{l}            &53.7 &\ul{54.7} &\textbf{55.2} &56.2 &\textbf{60.2} &\ul{57.1} \\
                                                       &\texttt{s}            &\ul{51.6} &49.5 &\textbf{53.7} &46.1 &\textbf{53.1} &\ul{47.3} \\
    \multirow[t]{2}{*}{\texttt{-}LogP}                 &\texttt{l}            &53.6 &\textbf{57.6} &\ul{55.2} &53.9 &\textbf{59.6} &\ul{58.0} \\
                                                       &\texttt{s}            &\ul{48.2} &\textbf{53.5} &47.3 &\ul{46.5} &\textbf{50.5} &44.9 \\
    \multirow[t]{2}{*}{\texttt{+}TPSA}                 &\texttt{l}            &48.8 &\textbf{56.5} &\ul{54.0} &\ul{55.1} &\textbf{56.8} &52.1 \\
                                                       &\texttt{s}            &49.5 &\ul{52.2} &\textbf{53.0} &\ul{51.2} &\textbf{56.7} &50.9 \\
    \multirow[t]{2}{*}{\texttt{-}TPSA}                 &\texttt{l}            &44.7 &\textbf{50.9} &\ul{48.3} &45.1 &\textbf{49.7} &\ul{45.6} \\
                                                       &\texttt{s}            &\ul{42.8} &\textbf{46.5} &40.7 &39.6 &\textbf{45.2} &\ul{40.2} \\
    \multirow[t]{2}{*}{\texttt{+}QED}                  &\texttt{l}            &48.8 &\textbf{56.6} &\ul{51.1} &47.6 &\textbf{56.6} &\ul{51.0} \\
                                                       &\texttt{s}            &46.3 &\textbf{55.3} &\ul{50.6} &42.1 &\textbf{56.6} &\ul{44.9} \\
    \multirow[t]{2}{*}{\texttt{-}QED}                  &\texttt{l}            &54.2 &\textbf{63.0} &\ul{55.5} &55.1 &\textbf{56.4} &\ul{56.2} \\
                                                       &\texttt{s}            &45.6 &\textbf{56.3} &\ul{47.5} &\ul{42.5} &\textbf{48.5} &40.1 \\
    \multirow[t]{2}{*}{Average}                        &\texttt{l}            &50.6 &\textbf{56.6} &\ul{53.2} &52.2 &\textbf{56.6} &\ul{53.3} \\
                                                       &\texttt{s}            &47.3 &\textbf{52.2} &\ul{48.8} &44.7 &\textbf{51.8} &\ul{44.7} \\
    \midrule
    \multirow[t]{2}{*}{\texttt{+}LogP \texttt{+}TPSA}  &\texttt{l} \texttt{l} &43.3 &\textbf{53.2} &\ul{44.9} &36.7 &\textbf{45.3} &\ul{41.0} \\
                                                       &\texttt{s} \texttt{s} &37.9 &\textbf{41.2} &\ul{40.3} &30.6 &\textbf{41.5} &\ul{36.8} \\
    \multirow[t]{2}{*}{\texttt{+}LogP \texttt{-}TPSA}  &\texttt{l} \texttt{l} &46.4 &\textbf{58.3} &\ul{47.6} &45.2 &\textbf{50.7} &\ul{46.0} \\
                                                       &\texttt{s} \texttt{s} &\ul{46.1} &\textbf{48.8} &43.6 &42.3 &\textbf{48.8} &\ul{43.5} \\
    \multirow[t]{2}{*}{\texttt{-}LogP \texttt{+}TPSA}  &\texttt{l} \texttt{l} &\ul{49.7} &\textbf{52.3} &46.5 &\ul{51.8} &\textbf{54.4} &50.6 \\
                                                       &\texttt{s} \texttt{s} &\ul{45.5} &\textbf{49.3} &43.4 &38.5 &\textbf{47.9} &\ul{40.1} \\
    \multirow[t]{2}{*}{\texttt{-}LogP \texttt{-}TPSA}  &\texttt{l} \texttt{l} &\ul{41.1} &\textbf{51.6} &40.9 &42.3 &\textbf{47.5} &\ul{44.5} \\
                                                       &\texttt{s} \texttt{s} &\ul{46.1} &\textbf{47.3} &44.9 &40.1 &\textbf{49.8} &\ul{41.8} \\
    \multirow[t]{2}{*}{\texttt{+}LogP \texttt{+}QED}   &\texttt{l} \texttt{l} &45.9 &\textbf{53.5} &\ul{47.7} &41.0 &\textbf{52.2} &\ul{43.7} \\
                                                       &\texttt{s} \texttt{s} &43.3 &\textbf{50.5} &\ul{48.2} &\ul{43.7} &\textbf{54.6} &42.6 \\
    \multirow[t]{2}{*}{\texttt{+}LogP \texttt{-}QED}   &\texttt{l} \texttt{l} &50.2 &\textbf{58.7} &\ul{50.6} &47.2 &\textbf{51.2} &\ul{49.1} \\
                                                       &\texttt{s} \texttt{s} &\ul{44.8} &\textbf{50.5} &43.3 &41.6 &\textbf{47.2} &\ul{45.1} \\
    \multirow[t]{2}{*}{\texttt{-}LogP \texttt{+}QED}   &\texttt{l} \texttt{l} &42.7 &\textbf{52.1} &\ul{46.3} &40.1 &\textbf{46.9} &\ul{42.0} \\
                                                       &\texttt{s} \texttt{s} &46.7 &\ul{50.3} &\textbf{52.8} &\ul{45.8} &\textbf{54.6} &45.3 \\
    \multirow[t]{2}{*}{\texttt{-}LogP \texttt{-}QED}   &\texttt{l} \texttt{l} &44.8 &\textbf{52.6} &\ul{46.4} &48.7 &\textbf{50.9} &\ul{50.7} \\
                                                       &\texttt{s} \texttt{s} &\textbf{47.0} &43.1 &\ul{44.0} &41.1 &\textbf{46.5} &\ul{42.9} \\
    \multirow[t]{2}{*}{Average}                        &\texttt{l} \texttt{l} &45.5 &\textbf{54.0} &\ul{46.4} &44.1 &\textbf{49.9} &\ul{46.0} \\
                                                       &\texttt{s} \texttt{s} &44.7 &\textbf{47.6} &\ul{45.1} &40.5 &\textbf{48.9} &\ul{42.3} \\
    \midrule
    %%%%%%%%%%%%%%%%%%%%%%%%%%%%%%%%%%%%%%%%%%%%%%%%%%%%%
    %%%%%%%%%%%%%%%%%%%%%%%%%%%%%%%%%%%%%%%%%%%%%%%%%%%%%
    \midrule
    \multirow[t]{3}{*}{$p$} &\multirow[t]{3}{*}{$d$} &\multicolumn{3}{c}{Llama-3.1-8B} &\multicolumn{3}{c}{Llama-3.1-70B} \\
    & &ChatDrug &AgentDrug &AgentDrug$\dag$ &ChatDrug &AgentDrug &AgentDrug$\dag$ \\
    & &similarity &similarity &similarity &similarity &similarity &similarity \\
    \midrule
    \multirow[t]{2}{*}{\texttt{+}LogP}                 &\texttt{l}            &44.1 &\textbf{47.3} &\ul{46.0} &68.7 &\textbf{70.7} &\ul{69.7} \\
                                                       &\texttt{s}            &37.9 &\ul{39.1} &\textbf{40.9} &51.4 &\textbf{54.9} &\ul{53.3} \\
    \multirow[t]{2}{*}{\texttt{-}LogP}                 &\texttt{l}            &38.8 &\textbf{46.2} &\ul{42.3} &69.0 &\textbf{70.9} &\ul{69.1} \\
                                                       &\texttt{s}            &33.2 &\ul{36.2} &\textbf{37.3} &51.6 &\textbf{56.5} &\ul{55.8} \\
    \multirow[t]{2}{*}{\texttt{+}TPSA}                 &\texttt{l}            &41.8 &\textbf{45.0} &\ul{44.9} &70.5 &\ul{70.8} &\textbf{71.0} \\
                                                       &\texttt{s}            &37.6 &\ul{39.1} &\textbf{40.6} &66.8 &\ul{67.8} &\textbf{68.1} \\
    \multirow[t]{2}{*}{\texttt{-}TPSA}                 &\texttt{l}            &\ul{37.7} &\textbf{38.0} &35.2 &52.7 &\ul{53.3} &\textbf{53.8} \\
                                                       &\texttt{s}            &\textbf{31.3} &\ul{31.0} &29.3 &\ul{45.2} &44.4 &\textbf{45.3} \\
    \multirow[t]{2}{*}{\texttt{+}QED}                  &\texttt{l}            &40.6 &\ul{41.0} &\textbf{41.4} &46.6 &\ul{51.5} &\textbf{52.6} \\
                                                       &\texttt{s}            &32.5 &\textbf{38.5} &\ul{34.1} &\ul{45.7} &\textbf{46.6} &44.4 \\
    \multirow[t]{2}{*}{\texttt{-}QED}                  &\texttt{l}            &\ul{46.7} &\textbf{47.1} &45.4 &64.0 &\ul{66.4} &\textbf{67.1} \\
                                                       &\texttt{s}            &33.4 &\textbf{34.1} &\ul{34.1} &40.2 &\ul{44.4} &\textbf{45.2} \\
    \multirow[t]{2}{*}{Average}                        &\texttt{l}            &41.6 &\textbf{44.1} &\ul{42.5} &61.9 &\textbf{63.9} &\ul{63.9} \\
                                                       &\texttt{s}            &34.3 &\textbf{36.3} &\ul{36.1} &50.2 &\textbf{52.4} &\ul{52.0} \\
    \midrule
    \multirow[t]{2}{*}{\texttt{+}LogP \texttt{+}TPSA}  &\texttt{l} \texttt{l} &\textbf{34.1} &\ul{32.9} &32.4 &39.7 &\ul{48.1} &\textbf{48.1} \\
                                                       &\texttt{s} \texttt{s} &\ul{29.0} &\textbf{32.3} &27.0 &29.7 &\ul{36.9} &\textbf{39.5} \\
    \multirow[t]{2}{*}{\texttt{+}LogP \texttt{-}TPSA}  &\texttt{l} \texttt{l} &\textbf{37.6} &35.5 &\ul{37.0} &49.5 &\textbf{51.9} &\ul{51.5} \\
                                                       &\texttt{s} \texttt{s} &31.1 &\ul{32.0} &\textbf{32.4} &41.9 &\textbf{45.2} &\ul{44.1} \\
    \multirow[t]{2}{*}{\texttt{-}LogP \texttt{+}TPSA}  &\texttt{l} \texttt{l} &35.3 &\textbf{38.0} &\ul{36.3} &65.0 &\textbf{69.0} &\ul{66.2} \\
                                                       &\texttt{s} \texttt{s} &\ul{31.7} &\textbf{34.0} &31.5 &52.9 &\textbf{55.3} &\ul{54.4} \\
    \multirow[t]{2}{*}{\texttt{-}LogP \texttt{-}TPSA}  &\texttt{l} \texttt{l} &33.3 &\ul{35.9} &\textbf{36.6} &41.4 &\textbf{47.6} &\ul{44.7} \\
                                                       &\texttt{s} \texttt{s} &\ul{32.8} &\textbf{34.3} &30.5 &36.8 &\ul{40.3} &\textbf{40.5} \\
    \multirow[t]{2}{*}{\texttt{+}LogP \texttt{+}QED}   &\texttt{l} \texttt{l} &36.1 &\ul{36.4} &\textbf{37.0} &45.0 &\ul{46.4} &\textbf{48.3} \\
                                                       &\texttt{s} \texttt{s} &\textbf{36.1} &\ul{35.4} &32.7 &42.5 &\ul{45.1} &\textbf{48.3} \\
    \multirow[t]{2}{*}{\texttt{+}LogP \texttt{-}QED}   &\texttt{l} \texttt{l} &31.5 &\ul{36.1} &\textbf{36.4} &57.1 &\textbf{59.6} &\ul{58.7} \\
                                                       &\texttt{s} \texttt{s} &20.8 &\ul{24.3} &\textbf{24.9} &42.3 &\textbf{48.8} &\ul{48.3} \\
    \multirow[t]{2}{*}{\texttt{-}LogP \texttt{+}QED}   &\texttt{l} \texttt{l} &\ul{37.7} &\textbf{40.1} &37.6 &42.5 &\textbf{52.9} &\ul{52.1} \\
                                                       &\texttt{s} \texttt{s} &32.0 &\textbf{36.2} &\ul{35.2} &44.9 &\textbf{51.4} &\ul{47.2} \\
    \multirow[t]{2}{*}{\texttt{-}LogP \texttt{-}QED}   &\texttt{l} \texttt{l} &36.0 &\textbf{41.6} &\ul{38.1} &58.5 &\ul{62.5} &\textbf{63.0} \\
                                                       &\texttt{s} \texttt{s} &\ul{21.4} &20.8 &\textbf{22.7} &41.5 &\textbf{47.2} &\ul{45.3} \\
    \multirow[t]{2}{*}{Average}                        &\texttt{l} \texttt{l} &35.2 &\textbf{37.1} &\ul{36.4} &49.8 &\textbf{54.8} &\ul{54.1} \\
                                                       &\texttt{s} \texttt{s} &29.4 &\textbf{31.2} &\ul{29.6} &41.6 &\textbf{46.3} &\ul{45.9} \\
    \midrule
    %%%%%%%%%%%%%%%%%%%%%%%%%%%%%%%%%%%%%%%%%%%%%%%%%%%%%

    %%%%%%%%%%%%%%%%%%%%%%%%%%%%%%%%%%%%%%%%%%%%%%%%%%%%%%%%%%%%%%
    \end{tabular*}
    \end{table}

Following previous work \cite{liu2024conversational}, we select 500 molecules from the ZINC database \cite{irwin2012zinc} as input molecules. Three molecular properties—LogP, TPSA, and QED—are used as editing targets. We evaluate performance on both single-property and multi-property editing tasks under two levels of difficulty: loose and strict thresholds, which define the required amounts of property increase or decrease (see Table~\ref{tab-A1} for threshold values).
We evaluate our results on four open-source LLMs: Qwen-2.5 \cite{yang2024qwen2} (3B and 7B) and LLaMA-3.1 \cite{dubey2024llama} (8B and 70B). As baselines, we include straightforward prompting (vanilla LLM response without iterative refinement) and a reinforcement learning-based method, REINVENT \cite{olivecrona2017molecular}.
More experiment settings are shown in the Appendix.

Tables~\ref{tab-2} and \ref{tab-3} report results on single- and multi-property editing tasks under both loose (\texttt{l}) and strict (\texttt{s}) thresholds, using three refinement iterations ($T = 3$).
First, thanks to its inner loop, AgentDrug significantly reduces the molecular hallucination, increasing the likelihood that the modified molecule is chemically valid. The performance gain of AgentDrug over its ablated version, AgentDrug$\dag$ (without the inner loop), directly demonstrates the effectiveness of this component.

The results show that AgentDrug consistently outperforms ChatDrug. With Qwen-2.5-3B, AgentDrug improves accuracy by 20.7\% (loose) and 16.8\% (strict) on six single-property objectives, and by 7.0\% and 5.3\% on eight multi-property objectives. With the larger Qwen-2.5-7B, the gains increase to 28.9\% and 29.0\% (single-property), and 14.9\% and 13.2\% (multi-property), under loose and strict thresholds, respectively.
Notably, AgentDrug$\dag$ also consistently outperforms ChatDrug, highlighting the value of the explicit \textit{gradient} signal, even without the inner loop.
While straightforward prompting achieves limited accuracy, the REINVENT baseline also fails to yield meaningful improvements—despite incurring high training costs. This is largely because REINVENT requires a supervised pretraining phase with labeled molecule pairs, which are often unavailable in practice.

\begin{figure}[!tbp]
\centering
\captionsetup[subfigure]{labelformat=empty}

\begin{subfigure}[b]{0.49\linewidth}
\includegraphics[width=\linewidth]{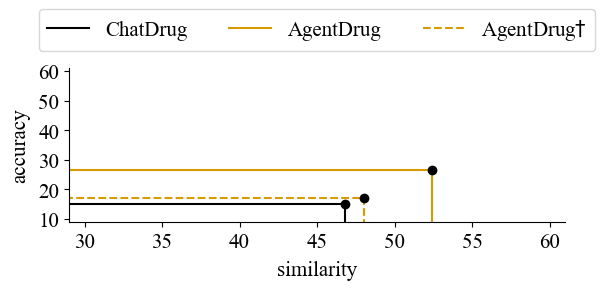}
\caption{Qwen2.5-3B}
\end{subfigure}%
\hfill
\begin{subfigure}[b]{0.49\linewidth}
\includegraphics[width=\linewidth]{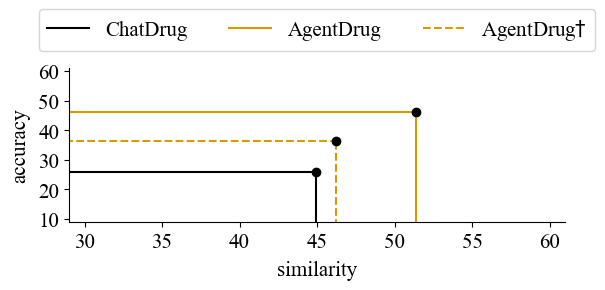}
\caption{Qwen2.5-7B}
\end{subfigure}%
\hfill
\begin{subfigure}[b]{0.49\linewidth}
\includegraphics[width=\linewidth]{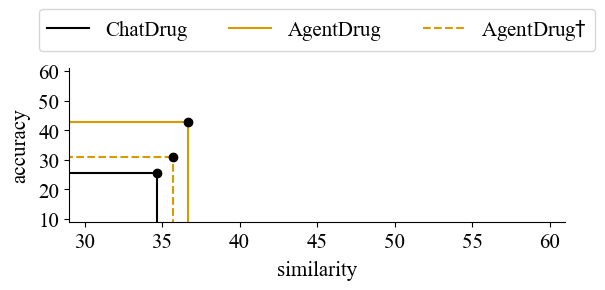}
\caption{Llama-3.1-8B}
\end{subfigure}%
\hfill
\begin{subfigure}[b]{0.49\linewidth}
\includegraphics[width=\linewidth]{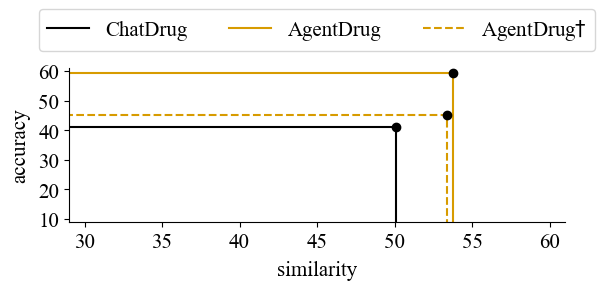}
\caption{Llama-3.1-70B}
\end{subfigure}%
\hfill

\caption{Comparison of accuracy vs. similarity trade-offs for different models. Each plot shows the performance of ChatDrug (black), AgentDrug (yellow solid), and AgentDrug$\dagger$ (yellow dashed) across four LLMs. 
}
\label{Figure-2}
\end{figure}

Beyond accuracy, AgentDrug is also more effective in preserving the similarity constraint. As illustrated in Figure~\ref{Figure-2}, both AgentDrug and AgentDrug$\dag$ explore molecular space more effectively than ChatDrug, achieving better editing while maintaining similarity, which further demonstrates the effectiveness of the \textit{gradient}-guided refinement.

\section{Ablation Studies}

\subsection{Impact of using the example molecule}
We conduct an ablation study to evaluate the impact of incorporating the retrieved example molecule $m_e$, a component that is overlooked in prior work \cite{liu2024conversational}. Table~\ref{tab-5} presents results for both AgentDrug and ChatDrug when $m_e$ is excluded. Even without $m_e$, AgentDrug/$m_e$ consistently outperforms ChatDrug/$m_e$, demonstrating the effectiveness of the remaining components in AgentDrug.
Using Qwen-2.5-3B, AgentDrug/$m_e$ improves accuracy by 11.2\% and 6.7\% under loose and strict thresholds, respectively, on six single-property objectives, and by 3.0\% and 1.4\% on eight multi-property objectives. With Qwen-2.5-7B, improvements are 10.1\% and 5.8\% (single-property), and 2.4\% and 0.9\% (multi-property), under loose and strict thresholds, respectively.

\begin{figure}[!t]
\centering
\captionsetup[subfigure]{labelformat=empty}

\begin{subfigure}[b]{00.49\linewidth}
\includegraphics[width=\linewidth]{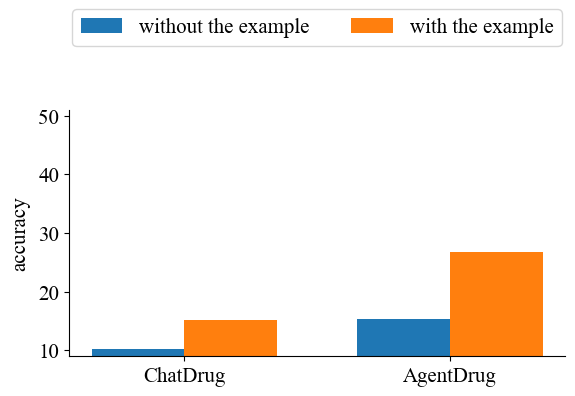}
\caption{Qwen2.5-3B}
\end{subfigure}%
\hfill
\begin{subfigure}[b]{00.49\linewidth}
\includegraphics[width=\linewidth]{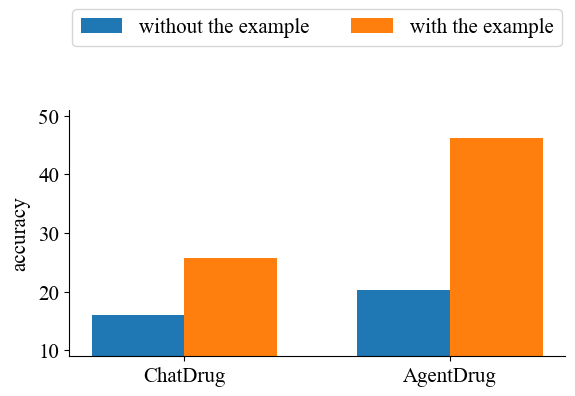}
\caption{Qwen2.5-7B}
\end{subfigure}%
\hfill

\caption{Impact of using the example molecule. Averaged results on all objectives. }
\label{Figure-3}
\end{figure}

Moreover, Figure~\ref{Figure-3} visually illustrates the contribution of incorporating the example molecule $m_e$ in both AgentDrug and ChatDrug. As discussed earlier, AgentDrug significantly reduces the molecular hallucination problem, increasing the likelihood that the modified molecule is valid. This, in turn, enables more reliable use of the retrieval step. By ensuring molecular validity, AgentDrug is able to fully leverage the retrieved example molecule $m_e$, thereby maximizing its effectiveness in guiding the editing process.

\subsection{Impact of the number of iterations}

We conduct an ablation study to examine how AgentDrug and ChatDrug behave under different numbers of refinement iterations $T$. Table~\ref{tab-6} presents results for both methods using $T = 4$, $5$, and $6$ iterations, respectively. The results show that AgentDrug consistently outperforms ChatDrug when using the same number of iterations. Notably, AgentDrug with fewer iterations still surpasses ChatDrug with more iterations, highlighting the superior efficiency and effectiveness of the AgentDrug framework.

\begin{figure}[!ht]
\centering
\captionsetup[subfigure]{labelformat=empty}

\begin{subfigure}[b]{00.49\linewidth}
\includegraphics[width=\linewidth]{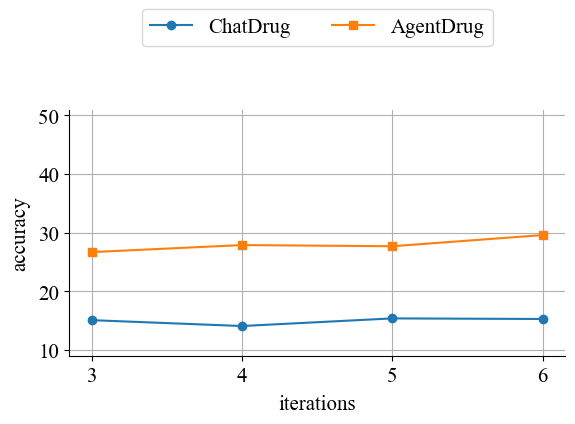}
\caption{Qwen2.5-3B}
\end{subfigure}%
\hfill
\begin{subfigure}[b]{00.49\linewidth}
\includegraphics[width=\linewidth]{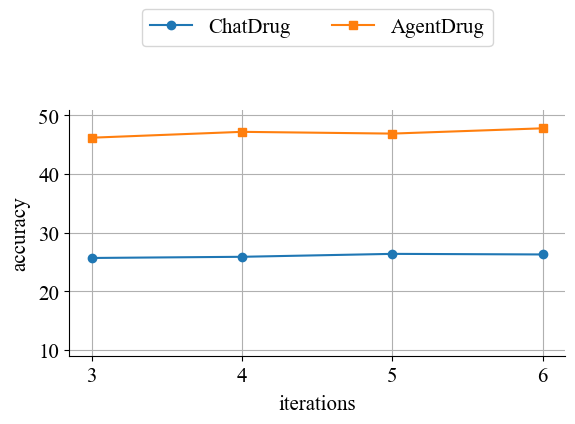}
\caption{Qwen2.5-7B}
\end{subfigure}%
\hfill

\caption{Impact of the number of iterations. Averaged results on all objectives. }
\label{Figure-4}
\end{figure}

However, as shown in Figure~\ref{Figure-4}, increasing the number of iterations does not lead to substantial performance gains for either AgentDrug or ChatDrug, indicating a saturation point in their effectiveness.

\section{Conclusion}
This paper introduced AgentDrug, a novel approach leveraging LLMs for zero-shot molecular editing through gradient-guided refinement. By providing explicit gradient information during the refinement process, AgentDrug enables targeted molecular modifications while maintaining similarity constraints. Our comprehensive experiments show that AgentDrug consistently outperforms powerful baselines on both single- and multi-property editing tasks. It is particularly effective for complex multi-property objectives and can balance well with editing accuracy and molecular similarity. 

\section*{Limitations}
First of all, although it does not require any training, we admit that utilizing LLMs in a loop of refinement requires prompting LLMs over multiple iterations, obviously leading to multiplying prompting costs \cite{samsi2023words}. However, this limitation is acceptable and compensated for by bringing improved performance. In addition, albeit containing a \textit{gradient} with directions and magnitudes to explicitly guide LLMs to refine the modified molecule toward the objective, feedback from AgentDrug currently lacks concrete actions towards the objective, which involve knowledge of molecular properties \cite{fang2023knowledge, hoang2024knowledge}, are presumably useful but hard to establish. 

\section{Acknowledgement}
This work was supported by the National Science Foundation (CHE–2202693) through the NSF Center for Computer Assisted Synthesis (C-CAS).

% \clearpage
\bibliography{refs}

\clearpage
\appendix

\section{Evaluation Settings}
Each editing objective is expressed in natural language using a standardized prompt template (see Box A1). 
Specifically, for each objective, we sample 1,000 molecules from the ZINC database and train LLMs using the REINFORCE algorithm \cite{sutton1999policy}, where the reward function is defined as $\prod_{i=1}^{N} \underset{p_i, d_i}{E} (\widehat{m})$. The models are fine-tuned using LoRA \cite{hu2022lora} with a rank of 16 and a learning rate of $1 \times 10^{-5}$, for a single training epoch.

\begin{table}[!ht]
\centering
\caption{The set of single- and multi-property objectives with loose and strict thresholds is used. }
\label{tab-A1}
\small
\setlength\tabcolsep{0pt}
\begin{tabular*}{\linewidth}{@{\extracolsep{\fill}} lll }

%%%%%%%%%%%%%%%%%%%%%%%%%%%%%%%%%%%%%%%%%%%%%%%%%%%%%
\midrule
$p$ &$d$ & \\
\midrule
\multirow[t]{2}{*}{\texttt{+}LogP}                 &\texttt{l}            &\texttt{+}00.0 \\
                                                   &\texttt{s}            &\texttt{+}00.5 \\
\multirow[t]{2}{*}{\texttt{-}LogP}                 &\texttt{l}            &\texttt{-}00.0 \\
                                                   &\texttt{s}            &\texttt{-}00.5 \\
\multirow[t]{2}{*}{\texttt{+}TPSA}                 &\texttt{l}            &\texttt{+}00.0 \\
                                                   &\texttt{s}            &\texttt{+}10.0 \\
\multirow[t]{2}{*}{\texttt{-}TPSA}                 &\texttt{l}            &\texttt{-}00.0 \\
                                                   &\texttt{s}            &\texttt{-}10.0 \\
\multirow[t]{2}{*}{\texttt{+}QED}                  &\texttt{l}            &\texttt{+}00.0 \\
                                                   &\texttt{s}            &\texttt{+}00.1 \\
\multirow[t]{2}{*}{\texttt{-}QED}                  &\texttt{l}            &\texttt{-}00.0 \\
                                                   &\texttt{s}            &\texttt{-}00.1 \\
\multirow[t]{2}{*}{Average}                        &\texttt{l}            &- \\
                                                   &\texttt{s}            &- \\
\midrule
\multirow[t]{2}{*}{\texttt{+}LogP \texttt{+}TPSA}  &\texttt{l} \texttt{l} &\texttt{+}00.0 \texttt{+}00.0 \\
                                                   &\texttt{s} \texttt{s} &\texttt{+}00.5 \texttt{+}10.0 \\
\multirow[t]{2}{*}{\texttt{+}LogP \texttt{-}TPSA}  &\texttt{l} \texttt{l} &\texttt{+}00.0 \texttt{-}00.0 \\
                                                   &\texttt{s} \texttt{s} &\texttt{+}00.5 \texttt{-}10.0 \\
\multirow[t]{2}{*}{\texttt{-}LogP \texttt{+}TPSA}  &\texttt{l} \texttt{l} &\texttt{-}00.0 \texttt{+}00.0 \\
                                                   &\texttt{s} \texttt{s} &\texttt{-}00.5 \texttt{+}10.0 \\
\multirow[t]{2}{*}{\texttt{-}LogP \texttt{-}TPSA}  &\texttt{l} \texttt{l} &\texttt{-}00.0 \texttt{-}00.0 \\
                                                   &\texttt{s} \texttt{s} &\texttt{-}00.5 \texttt{-}10.0 \\
\multirow[t]{2}{*}{\texttt{+}LogP \texttt{+}QED}   &\texttt{l} \texttt{l} &\texttt{+}00.0 \texttt{+}00.0 \\
                                                   &\texttt{s} \texttt{s} &\texttt{+}00.5 \texttt{+}00.1 \\
\multirow[t]{2}{*}{\texttt{+}LogP \texttt{-}QED}   &\texttt{l} \texttt{l} &\texttt{+}00.0 \texttt{-}00.0 \\
                                                   &\texttt{s} \texttt{s} &\texttt{+}00.5 \texttt{-}00.1 \\
\multirow[t]{2}{*}{\texttt{-}LogP \texttt{+}QED}   &\texttt{l} \texttt{l} &\texttt{-}00.0 \texttt{+}00.0 \\
                                                   &\texttt{s} \texttt{s} &\texttt{-}00.5 \texttt{+}00.1 \\
\multirow[t]{2}{*}{\texttt{-}LogP \texttt{-}QED}   &\texttt{l} \texttt{l} &\texttt{-}00.0 \texttt{-}00.0 \\
                                                   &\texttt{s} \texttt{s} &\texttt{-}00.5 \texttt{-}00.1 \\
\multirow[t]{2}{*}{Average}                        &\texttt{l} \texttt{l} &- \\
                                                   &\texttt{s} \texttt{s} &- \\
\midrule
%%%%%%%%%%%%%%%%%%%%%%%%%%%%%%%%%%%%%%%%%%%%%%%%%%%%%

\end{tabular*}
\end{table}

\begin{tcolorbox}[
left=5pt, right=5pt, top=5pt, bottom=5pt, fonttitle=\normalsize\bfseries, colback=gray!10, colframe=gray!40, title={
Box A1: Template for wrapping an objective in natural language to form the prompt
}]

\begin{minipage}{\textwidth}
\normalsize
Given [a given molecule $m$], modify it to [increase or decrease] its [desired properties $\{p_i\}_{i=1}^N$] by [thresholds $\{|d_i|\}_{i=1}^N$], respectively. Importantly, the modified molecule must be similar to the given one.

\vspace{0.5em}
Respond with only the SMILES string of the modified molecule. No explanation is needed.
\end{minipage}
\end{tcolorbox}

\section{Ablation Studies}

\begin{table}[!t]
    \centering
    \caption{The results (T = 3) on a set of single- and multi-property objectives with loose and strict thresholds. }
    \label{tab-5}
    \tiny
    \setlength\tabcolsep{1.5pt}
    \begin{tabular*}{\linewidth}{@{\extracolsep{\fill}} llcccc }
    %%%%%%%%%%%%%%%%%%%%%%%%%%%%%%%%%%%%%%%%%%%%%%%%%%%%%%%%%%%%%%

    %%%%%%%%%%%%%%%%%%%%%%%%%%%%%%%%%%%%%%%%%%%%%%%%%%%%%
    \midrule
    \multirow[t]{3}{*}{$p$} &\multirow[t]{3}{*}{$d$} &\multicolumn{2}{c}{Qwen2.5-3B} &\multicolumn{2}{c}{Qwen2.5-7B} \\
    & &ChatDrug &AgentDrug &ChatDrug &AgentDrug \\
    & &accuracy &accuracy &accuracy &accuracy \\
    \midrule
    \multirow[t]{2}{*}{\texttt{+}LogP}                 &\texttt{l}            &25.3 &\textbf{39.3} &26.1 &\textbf{49.0} \\
                                                       &\texttt{s}            &18.5 &\textbf{28.8} &18.1 &\textbf{28.6} \\
    \multirow[t]{2}{*}{\texttt{-}LogP}                 &\texttt{l}            &21.6 &\textbf{27.7} &39.7 &\textbf{45.9} \\
                                                       &\texttt{s}            &09.7 &\textbf{15.7} &24.8 &\textbf{25.8} \\
    \multirow[t]{2}{*}{\texttt{+}TPSA}                 &\texttt{l}            &16.8 &\textbf{28.7} &24.7 &\textbf{36.5} \\
                                                       &\texttt{s}            &15.8 &\textbf{29.5} &22.7 &\textbf{35.1} \\
    \multirow[t]{2}{*}{\texttt{-}TPSA}                 &\texttt{l}            &14.5 &\textbf{22.0} &\textbf{20.4} &18.1 \\
                                                       &\texttt{s}            &\textbf{13.9} &13.1 &\textbf{12.3} &10.8 \\
    \multirow[t]{2}{*}{\texttt{+}QED}                  &\texttt{l}            &19.6 &\textbf{25.2} &23.7 &\textbf{29.5} \\
                                                       &\texttt{s}            &05.6 &\textbf{06.3} &\textbf{05.9} &05.2 \\
    \multirow[t]{2}{*}{\texttt{-}QED}                  &\texttt{l}            &21.6 &\textbf{44.0} &38.4 &\textbf{54.2} \\
                                                       &\texttt{s}            &07.9 &\textbf{18.4} &16.3 &\textbf{29.3} \\
    \multirow[t]{2}{*}{Average}                        &\texttt{l}            &19.9 &\textbf{31.1} &28.8 &\textbf{38.9} \\
                                                       &\texttt{s}            &11.9 &\textbf{18.6} &16.7 &\textbf{22.5} \\
    \midrule
    \multirow[t]{2}{*}{\texttt{+}LogP \texttt{+}TPSA}  &\texttt{l} \texttt{l} &06.5 &\textbf{10.2} &08.0 &\textbf{11.2} \\
                                                        &\texttt{s} \texttt{s} &01.6 &\textbf{07.4} &05.6 &\textbf{09.6} \\
    \multirow[t]{2}{*}{\texttt{+}LogP \texttt{-}TPSA}  &\texttt{l} \texttt{l} &07.2 &\textbf{11.6} &\textbf{11.8} &09.6 \\
                                                        &\texttt{s} \texttt{s} &\textbf{06.2} &05.3 &\textbf{08.1} &03.6 \\
    \multirow[t]{2}{*}{\texttt{-}LogP \texttt{+}TPSA}  &\texttt{l} \texttt{l} &09.3 &\textbf{15.8} &24.6 &\textbf{35.2} \\
                                                        &\texttt{s} \texttt{s} &03.8 &\textbf{08.4} &13.3 &\textbf{21.8} \\
    \multirow[t]{2}{*}{\texttt{-}LogP \texttt{-}TPSA}  &\texttt{l} \texttt{l} &\textbf{06.3} &05.0 &\textbf{08.0} &02.9 \\
                                                        &\texttt{s} \texttt{s} &\textbf{01.3} &00.7 &\textbf{02.8} &01.1 \\
    \multirow[t]{2}{*}{\texttt{+}LogP \texttt{+}QED}   &\texttt{l} \texttt{l} &09.6 &\textbf{12.3} &12.2 &\textbf{13.6} \\
                                                        &\texttt{s} \texttt{s} &\textbf{02.2} &01.6 &01.3 &\textbf{01.8} \\
    \multirow[t]{2}{*}{\texttt{+}LogP \texttt{-}QED}   &\texttt{l} \texttt{l} &13.3 &\textbf{23.3} &21.9 &\textbf{27.7} \\
                                                        &\texttt{s} \texttt{s} &05.8 &\textbf{07.6} &\textbf{08.0} &07.9 \\
    \multirow[t]{2}{*}{\texttt{-}LogP \texttt{+}QED}   &\texttt{l} \texttt{l} &08.3 &\textbf{09.1} &\textbf{14.2} &13.5 \\
                                                        &\texttt{s} \texttt{s} &\textbf{01.3} &00.7 &01.4 &\textbf{01.5} \\
    \multirow[t]{2}{*}{\texttt{-}LogP \texttt{-}QED}   &\texttt{l} \texttt{l} &\textbf{11.2} &09.0 &25.0 &\textbf{31.2} \\
                                                        &\texttt{s} \texttt{s} &03.0 &\textbf{04.4} &07.8 &\textbf{08.7} \\
    \multirow[t]{2}{*}{Average}                        &\texttt{l} \texttt{l} &09.0 &\textbf{12.0} &15.7 &\textbf{18.1} \\
                                                        &\texttt{s} \texttt{s} &03.1 &\textbf{04.5} &06.1 &\textbf{07.0} \\
    \midrule
    %%%%%%%%%%%%%%%%%%%%%%%%%%%%%%%%%%%%%%%%%%%%%%%%%%%%%

    %%%%%%%%%%%%%%%%%%%%%%%%%%%%%%%%%%%%%%%%%%%%%%%%%%%%%%%%%%%%%%
    \end{tabular*}
    \end{table}

\begin{table*}[!t]
    \centering
    \caption{The results (T = 4, 5, 6) on a set of single- and multi-property objectives with loose and strict thresholds. }
    \label{tab-6}
    \tiny
    \setlength\tabcolsep{0pt}
    \begin{tabular*}{\linewidth}{@{\extracolsep{\fill}} llcccccccccccccccccccc }
    %%%%%%%%%%%%%%%%%%%%%%%%%%%%%%%%%%%%%%%%%%%%%%%%%%%%%%%%%%%%%%%%%%%%%%%%%%%%

    %%%%%%%%%%%%%%%%%%%%%%%%%%%%%%%%%%%%%%%%%%%%%%%%%%%%%
    \midrule
    \multirow[t]{3}{*}{$p$} &\multirow[t]{3}{*}{$d$} &\multicolumn{6}{c}{Qwen2.5-3B} &\multicolumn{6}{c}{Qwen2.5-7B} \\
    & &ChatDrug-4 &AgentDrug-4 &ChatDrug-5 &AgentDrug-5 &ChatDrug-6 &AgentDrug-6 &ChatDrug-4 &AgentDrug-4 &ChatDrug-5 &AgentDrug-5 &ChatDrug-6 &AgentDrug-6 \\
    & &accuracy &accuracy &accuracy &accuracy &accuracy &accuracy &accuracy &accuracy &accuracy &accuracy &accuracy &accuracy \\
    \midrule
    \multirow[t]{2}{*}{\texttt{+}LogP}                 &\texttt{l}            &28.2 &\textbf{53.5} &28.1 &\textbf{49.6} &31.5 &\textbf{54.5} &45.2 &\textbf{68.8} &43.8 &\textbf{71.5} &41.6 &\textbf{73.0} \\
                                                       &\texttt{s}            &20.4 &\textbf{41.5} &25.8 &\textbf{41.5} &22.4 &\textbf{47.5} &32.1 &\textbf{65.0} &34.3 &\textbf{65.0} &32.1 &\textbf{72.2} \\
    \multirow[t]{2}{*}{\texttt{-}LogP}                 &\texttt{l}            &19.9 &\textbf{50.6} &26.1 &\textbf{45.6} &25.5 &\textbf{53.8} &40.9 &\textbf{70.4} &41.9 &\textbf{70.8} &38.2 &\textbf{76.4} \\
                                                       &\texttt{s}            &18.3 &\textbf{38.0} &20.6 &\textbf{38.1} &19.9 &\textbf{47.0} &39.6 &\textbf{71.4} &38.1 &\textbf{73.8} &39.9 &\textbf{71.4} \\
    \multirow[t]{2}{*}{\texttt{+}TPSA}                 &\texttt{l}            &21.6 &\textbf{47.7} &23.9 &\textbf{48.3} &21.4 &\textbf{49.6} &36.1 &\textbf{70.5} &37.1 &\textbf{58.6} &40.0 &\textbf{62.9} \\
                                                       &\texttt{s}            &21.9 &\textbf{41.5} &23.6 &\textbf{44.1} &26.1 &\textbf{48.5} &33.4 &\textbf{53.1} &27.9 &\textbf{52.9} &29.6 &\textbf{55.3} \\
    \multirow[t]{2}{*}{\texttt{-}TPSA}                 &\texttt{l}            &21.8 &\textbf{41.5} &24.2 &\textbf{41.1} &25.6 &\textbf{45.0} &36.0 &\textbf{66.8} &33.0 &\textbf{72.8} &39.5 &\textbf{72.6} \\
                                                       &\texttt{s}            &24.3 &\textbf{41.8} &21.5 &\textbf{40.3} &27.4 &\textbf{45.5} &30.8 &\textbf{58.4} &36.9 &\textbf{71.0} &38.9 &\textbf{66.4} \\
    \multirow[t]{2}{*}{\texttt{+}QED}                  &\texttt{l}            &22.6 &\textbf{44.8} &24.7 &\textbf{48.6} &23.8 &\textbf{55.1} &34.2 &\textbf{71.8} &40.1 &\textbf{69.6} &39.0 &\textbf{77.2} \\
                                                       &\texttt{s}            &08.0 &\textbf{20.2} &13.3 &\textbf{23.2} &11.3 &\textbf{25.0} &22.2 &\textbf{53.5} &26.6 &\textbf{51.0} &20.7 &\textbf{47.8} \\
    \multirow[t]{2}{*}{\texttt{-}QED}                  &\texttt{l}            &28.5 &\textbf{61.9} &34.0 &\textbf{63.4} &29.6 &\textbf{54.8} &44.6 &\textbf{72.4} &45.9 &\textbf{81.3} &46.3 &\textbf{74.2} \\
                                                       &\texttt{s}            &13.1 &\textbf{44.0} &17.2 &\textbf{44.2} &15.8 &\textbf{50.0} &33.8 &\textbf{64.7} &38.2 &\textbf{64.9} &33.8 &\textbf{64.9} \\
    \multirow[t]{2}{*}{Average}                        &\texttt{l}            &23.8 &\textbf{50.0} &26.8 &\textbf{49.4} &26.2 &\textbf{52.1} &39.5 &\textbf{70.1} &40.3 &\textbf{70.8} &40.8 &\textbf{72.7} \\
                                                       &\texttt{s}            &17.7 &\textbf{37.8} &20.3 &\textbf{38.6} &20.5 &\textbf{43.9} &32.0 &\textbf{61.0} &33.7 &\textbf{63.1} &32.5 &\textbf{63.0} \\
    \midrule
    \multirow[t]{2}{*}{\texttt{+}LogP \texttt{+}TPSA}  &\texttt{l} \texttt{l} &15.4 &\textbf{24.7} &15.4 &\textbf{25.0} &15.4 &\textbf{27.1} &26.8 &\textbf{60.8} &26.8 &\textbf{51.4} &26.8 &\textbf{52.8} \\
                                                       &\texttt{s} \texttt{s} &16.9 &\textbf{22.7} &16.9 &\textbf{23.3} &16.9 &\textbf{25.5} &22.0 &\textbf{50.2} &22.0 &\textbf{46.1} &22.0 &\textbf{49.4} \\
    \multirow[t]{2}{*}{\texttt{+}LogP \texttt{-}TPSA}  &\texttt{l} \texttt{l} &08.8 &\textbf{15.7} &08.8 &\textbf{12.8} &08.8 &\textbf{14.0} &18.6 &\textbf{27.7} &18.6 &\textbf{27.1} &18.6 &\textbf{30.6} \\
                                                       &\texttt{s} \texttt{s} &03.5 &\textbf{07.9} &03.5 &\textbf{06.9} &03.5 &\textbf{06.9} &10.6 &\textbf{16.5} &10.6 &\textbf{14.1} &10.6 &\textbf{18.1} \\
    \multirow[t]{2}{*}{\texttt{-}LogP \texttt{+}TPSA}  &\texttt{l} \texttt{l} &15.7 &\textbf{22.6} &15.7 &\textbf{24.5} &15.7 &\textbf{22.6} &39.6 &\textbf{53.8} &39.6 &\textbf{56.0} &39.6 &\textbf{58.0} \\
                                                       &\texttt{s} \texttt{s} &10.1 &\textbf{22.8} &10.1 &\textbf{24.7} &10.1 &\textbf{24.7} &25.8 &\textbf{50.2} &25.8 &\textbf{54.2} &25.8 &\textbf{54.2} \\
    \multirow[t]{2}{*}{\texttt{-}LogP \texttt{-}TPSA}  &\texttt{l} \texttt{l} &05.1 &\textbf{06.2} &05.1 &\textbf{08.2} &05.1 &\textbf{08.8} &05.1 &\textbf{12.6} &05.1 &\textbf{12.8} &05.1 &\textbf{12.4} \\
                                                       &\texttt{s} \texttt{s} &01.3 &\textbf{01.4} &01.3 &\textbf{02.8} &01.3 &\textbf{01.7} &02.3 &\textbf{05.8} &02.3 &\textbf{06.4} &02.3 &\textbf{05.5} \\
    \multirow[t]{2}{*}{\texttt{+}LogP \texttt{+}QED}   &\texttt{l} \texttt{l} &15.7 &\textbf{32.1} &15.7 &\textbf{26.7} &15.7 &\textbf{24.4} &31.0 &\textbf{61.9} &31.0 &\textbf{56.7} &31.0 &\textbf{57.5} \\
                                                       &\texttt{s} \texttt{s} &07.3 &\textbf{12.7} &07.3 &\textbf{14.8} &07.3 &\textbf{14.9} &14.0 &\textbf{38.8} &14.0 &\textbf{37.7} &14.0 &\textbf{38.2} \\
    \multirow[t]{2}{*}{\texttt{+}LogP \texttt{-}QED}   &\texttt{l} \texttt{l} &14.4 &\textbf{22.0} &14.4 &\textbf{24.4} &14.4 &\textbf{21.6} &20.3 &\textbf{21.5} &20.3 &\textbf{22.5} &20.3 &\textbf{23.3} \\
                                                       &\texttt{s} \texttt{s} &03.9 &\textbf{10.0} &03.9 &\textbf{08.6} &03.9 &\textbf{08.7} &03.9 &\textbf{06.8} &\textbf{03.9} &03.8 &03.9 &\textbf{07.9} \\
    \multirow[t]{2}{*}{\texttt{-}LogP \texttt{+}QED}   &\texttt{l} \texttt{l} &11.8 &\textbf{28.8} &11.8 &\textbf{23.8} &11.8 &\textbf{29.2} &33.0 &\textbf{59.0} &33.0 &\textbf{56.5} &33.0 &\textbf{54.9} \\
                                                       &\texttt{s} \texttt{s} &05.6 &\textbf{10.1} &05.6 &\textbf{08.0} &05.6 &\textbf{08.9} &17.9 &\textbf{34.2} &17.9 &\textbf{35.7} &17.9 &\textbf{32.7} \\
    \multirow[t]{2}{*}{\texttt{-}LogP \texttt{-}QED}   &\texttt{l} \texttt{l} &07.8 &\textbf{12.2} &07.8 &\textbf{11.3} &07.8 &\textbf{11.6} &19.7 &\textbf{28.4} &19.7 &\textbf{23.3} &19.7 &\textbf{22.5} \\
                                                       &\texttt{s} \texttt{s} &\textbf{03.8} &02.8 &\textbf{03.8} &02.8 &\textbf{03.8} &02.5 &05.6 &\textbf{05.8} &05.6 &\textbf{05.9} &\textbf{05.6} &05.3 \\
    \multirow[t]{2}{*}{Average}                        &\texttt{l} \texttt{l} &11.8 &\textbf{20.5} &11.8 &\textbf{19.6} &11.8 &\textbf{19.9} &24.3 &\textbf{40.7} &24.3 &\textbf{38.3} &24.3 &\textbf{39.0} \\
                                                       &\texttt{s} \texttt{s} &06.5 &\textbf{11.3} &06.5 &\textbf{11.5} &06.5 &\textbf{11.7} &12.8 &\textbf{26.0} &12.8 &\textbf{25.5} &12.8 &\textbf{26.4} \\
    \midrule
    %%%%%%%%%%%%%%%%%%%%%%%%%%%%%%%%%%%%%%%%%%%%%%%%%%%%%

    %%%%%%%%%%%%%%%%%%%%%%%%%%%%%%%%%%%%%%%%%%%%%%%%%%%%%%%%%%%%%%%%%%%%%%%%%%%%
    \end{tabular*}
    \end{table*}

\subsection{Impact of using the example molecule}

Table~\ref{tab-5} presents results for both AgentDrug and ChatDrug when $m_e$ is excluded. 

\subsection{Impact of the number of iterations}

Table~\ref{tab-6} presents results for both methods using $T = 4$, $5$, and $6$ iterations, respectively. 

\section{Related Work}

Recent advancements in AI, particularly in large language models (LLMs), have begun to transform the drug discovery process by efficiently scaling many core molecular tasks \cite{mak2019artificial}. Notably, \citet{seidl2023enhancing, zhao2023gimlet, liu-etal-2024-reactxt} have incrementally pre-trained LLMs to advance molecular property prediction, synthesis, and retrosynthesis. In addition, \citet{edwards-etal-2022-translation, liu-etal-2023-molca, li2024empowering, le2024molx} have trained LLMs for molecular captioning. However, molecular editing, which plays a critical role in designing improved compounds, remains underexplored.

Traditionally, supervised learning has been the dominant approach for addressing this task. For example, \citet{he2021molecular, he2022transformer} construct datasets containing millions of labeled molecule pairs and train models from scratch. Building on this foundation, \citet{wu2024leveraging} incorporate external atomic embeddings into model training, while \citet{turutov2024molecular} introduce a patentability penalty to guide the learning process. Additionally, \citet{nahal2024human} propose a framework that involves chemists directly in the training loop to improve model alignment with expert knowledge. In contrast to these supervised approaches, recent work explores the use of large language models (LLMs) to bypass the need for labeled molecule pairs \cite{liu2023pre}. LLMs offer strong zero-shot and open-vocabulary generalization capabilities, allowing them to operate beyond fixed objective sets. As a result, they present a promising alternative for molecular editing while substantially reducing the cost and complexity of training.

As early pioneers, \citet{zhang2024future} investigate the use of LLMs for molecular editing through straightforward prompting, but observe limited accuracy. Inspired by the refinement strategies commonly employed by chemists, \citet{liu2024conversational} propose ChatDrug, a simple chat-based framework that guides LLMs through a refinement loop. In this loop, the LLM receives generic feedback such as ``The modified molecule does not meet the objective'' to revise the output toward the desired properties. 
However, relying solely on generic feedback leads the LLM to refine molecules in a largely unguided manner. Moreover, ChatDrug does not address the issue of molecular hallucination \cite{guo2023can}, where the LLM generates chemically invalid structures. As a result, the retrieval step is frequently disabled due to the high frequency of invalid outputs, limiting its overall effectiveness.

\section{Discussion}
In this section, we give a discussion on the motivation and nature of molecular editing. Despite the existence of huge chemical databases like ZINC, ChEMBL, PubChem, or commercial catalogs containing billions of molecules, design and editing are still essential for several reasons:
\begin{itemize}[noitemsep]
    \item The estimated drug-like chemical space is > $10^{60}$ molecules. Even huge databases like ZINC, ChEMBL, and PubChem are just a small fraction of that space. Thus, the chance that the target molecule (satisfying multiple objectives) is already enumerated is low. Moreover, retrieval over huge molecule databases is extremely expensive because computing similarity between molecules is hard to parallelize.
    \item Novelty for patentability. Pharmaceutical companies need novel chemical compounds they can patent. Retrieving a molecule already reported in the literature often means that they cannot claim exclusivity, making it commercially unviable.
\end{itemize}

In summary, retrieval from databases is only a valuable starting point to augment design and editing. For example, AgentDrug leverages a small database (10K molecules) to augment the editing process. It depends on the size of the database; the molecule $m_e$ can be easier or harder to find. Also, the retrieval may fail or return a low-similarity molecule in sparse areas.

\section{Full Algorithm}

\begin{algorithm}[!ht]
\caption{AgentDrug}
\label{Algorithm-1}
\scriptsize
\KwData{a given molecule $m$, an LLM $\mathcal{M}$. }
$\widehat{m} = \mathcal{M}(p || d || m)$ \hspace*{\fill} $\triangleright$ initiated\\
\For{\upshape iteration \textbf{in} 1, ..., T}{
    $\text{is}\_\text{valid}, \text{ParseError} = \mathcal{T}(\widehat{m}, \text{parsing})$ \\
    \If{\upshape $\text{is}\_\text{valid} = 0$}{
        $\widehat{m} = \mathcal{M}(p || d || m || \widehat{m} || \text{ParseError})$ \\
    }
    \Else{
        \If{\upshape {$\prod_{i=1}^{N} \underset{p_i, d_i}{E} (\widehat{m}) = 1$}}{
            \textbf{break}
        }
        \Else{
            $\widehat{m} = \mathcal{M}(p || d || m || \widehat{m} || \nabla_p || m_e)$
        }
    }
}
\textbf{Return} $\widehat{m}$
\end{algorithm}

\section{Examples}

\begin{figure*}[t]
\centering
\begin{subfigure}[t]{\textwidth}
\centering
\includegraphics[width=\textwidth]{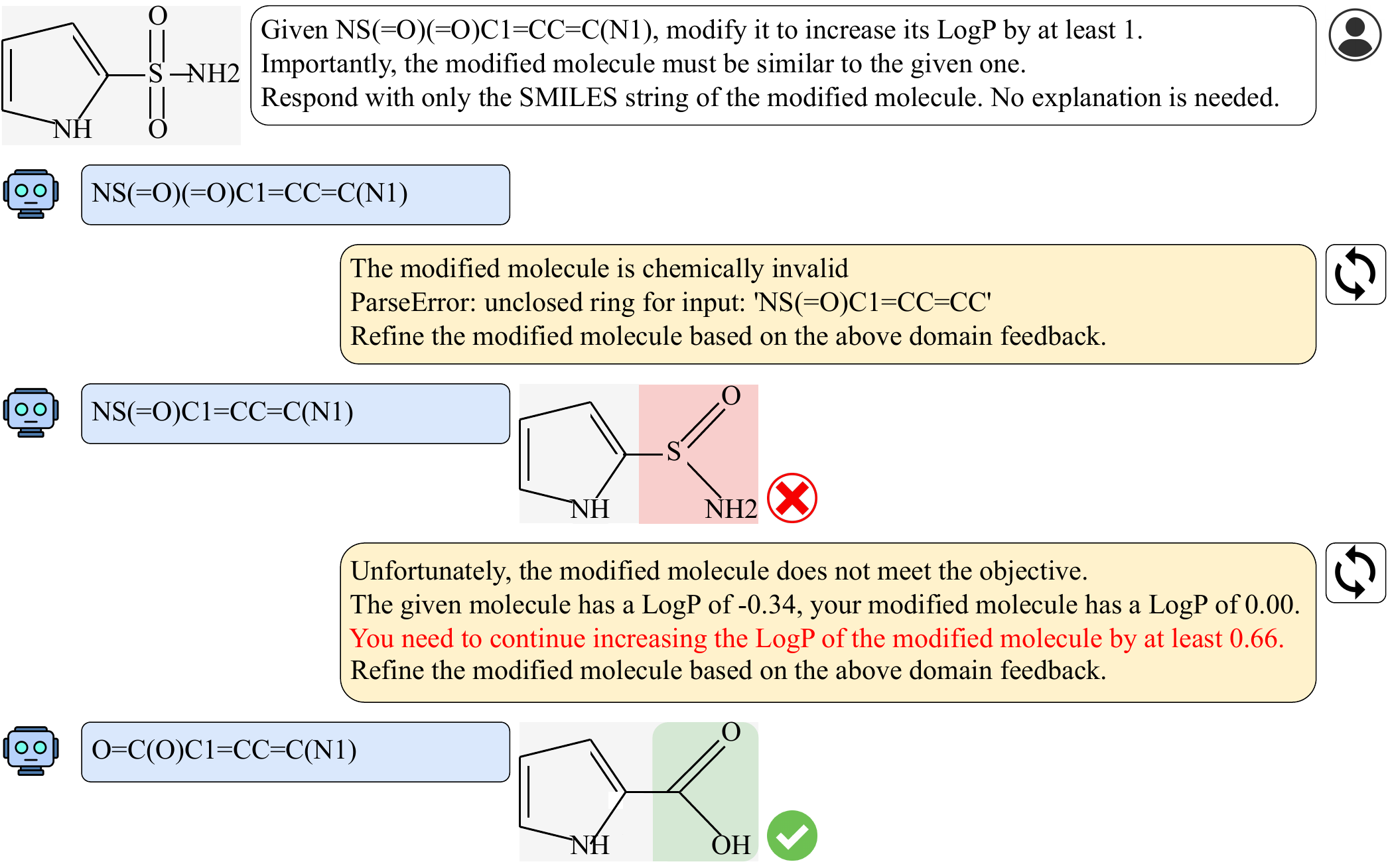}
\end{subfigure}

% \vspace{1.5em}

\begin{subfigure}[t]{\textwidth}
\centering
\includegraphics[width=\textwidth]{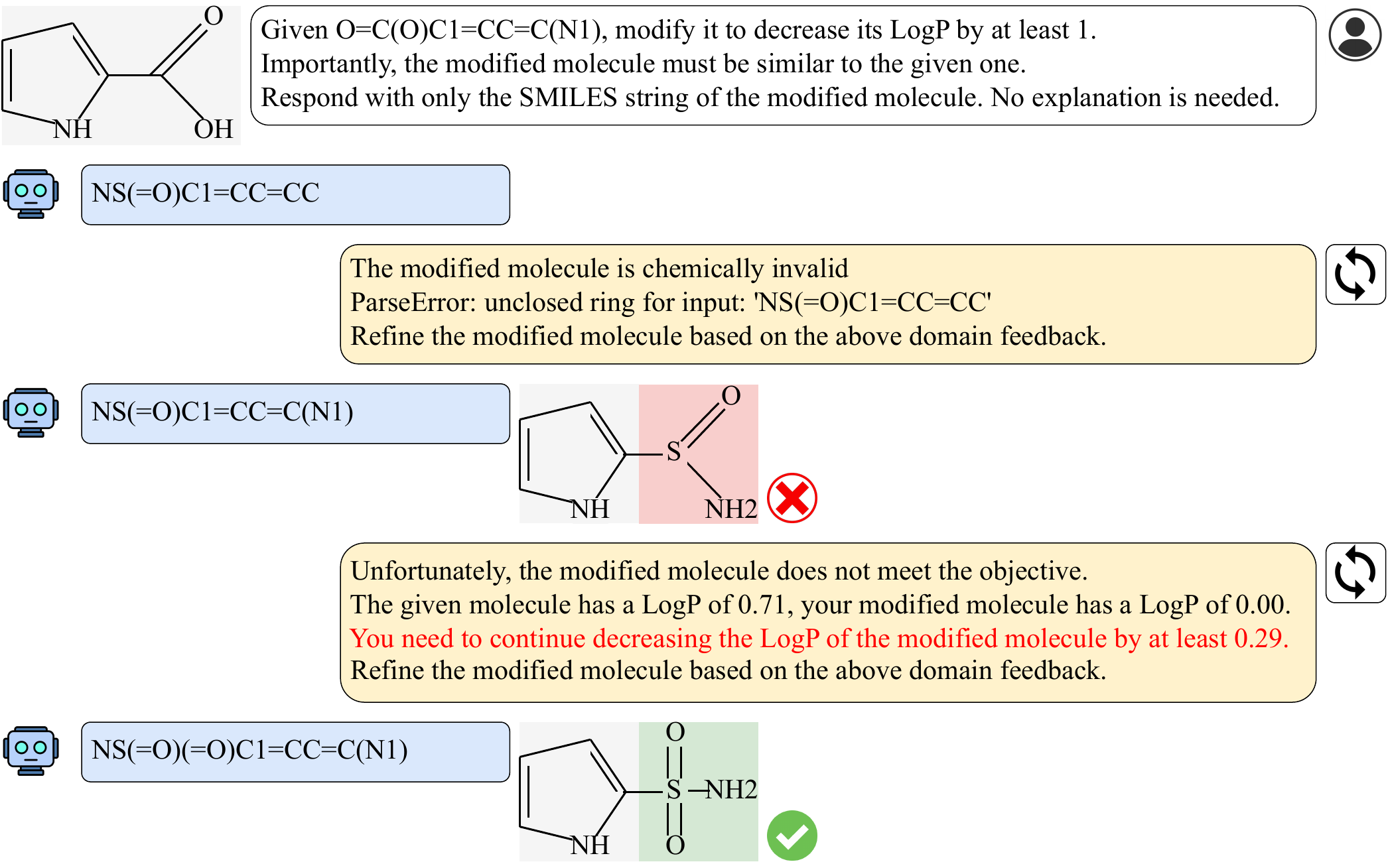}
\end{subfigure}

\caption{Illustrative examples of how AgentDrug works, including detailed prompts. }
\label{Figure-5}
\end{figure*}

Illustrative examples of AgentDrug’s workflow, including detailed prompts, are provided in Figure~\ref{Figure-5}. The \textit{gradient} signal used to guide the LLM is highlighted in \textcolor{red}{red} for clarity. 

\end{document}